%% file: vggface2_camera_ready.tex
\documentclass[10pt, conference, compsocconf]{IEEEtran}
\usepackage{cite}

\ifCLASSINFOpdf
  \usepackage[pdftex]{graphicx}
\else
\fi

\usepackage[tight,footnotesize]{subfigure}
\usepackage{caption}
\usepackage[pagebackref=true,breaklinks=true,letterpaper=true,colorlinks,bookmarks=false]{hyperref}
\usepackage{epsfig}
\usepackage{amsmath}
\usepackage{amssymb}
\begin{document}
\IEEEoverridecommandlockouts\pubid{\makebox[\columnwidth]{978-1-5386-2335-0/18/\$31.00~\copyright{}2018 IEEE \hfill}
\hspace{\columnsep}\makebox[\columnwidth]{ }}

\newcommand{\NewVGGFace}{VGGFace2}
\newcommand{\shortNewVGGFace}{VF2}
\newcommand{\CNNs}{Convolutional Neural Networks}
\renewcommand{\baselinestretch}{0.98}
\renewcommand{\thepage}{}
%
\title{\NewVGGFace{}: A dataset for recognising faces across pose and age}

\author{\parbox{12cm}{\centering
    {Qiong Cao, Li Shen, Weidi Xie, Omkar M.\ Parkhi and Andrew Zisserman}\\
    {\normalsize Visual Geometry Group, Department of Engineering Science, University of Oxford}\\
    {\normalsize \{qiong,lishen,weidi,omkar,az\}@robots.ox.ac.uk}}
}

\maketitle

\begin{abstract}
In this paper, 
we introduce a new large-scale face dataset named \NewVGGFace{}.
The dataset contains $3.31$ million images of $9131$ subjects, 
with an average of $362.6$ images for each subject.  
Images are downloaded from Google Image Search and have large variations in pose, age, illumination,
ethnicity and profession (e.g.\ actors, athletes, politicians). 

The dataset was collected with three goals in mind: 
(i) to have both a large number of identities and also a large number of images for each identity; 
(ii) to cover a large range of pose, age and ethnicity; and
(iii) to minimise the label noise.  
We describe how the dataset was collected, in particular the automated and manual filtering stages to
ensure a high accuracy for the images of each identity.

To assess face recognition performance using the new dataset, we train ResNet-50 (with and without Squeeze-and-Excitation blocks)  \CNNs{} on \NewVGGFace{}, 
on MS-Celeb-1M, and on their union,
and show that training on \NewVGGFace{} leads to improved recognition performance over pose and age.
Finally, using the models trained on these datasets,
we demonstrate state-of-the-art performance on the face recognition of IJB datasets, 
exceeding the previous state-of-the-art by a large margin.
The dataset and models are publicly available\footnote{\url{http://www.robots.ox.ac.uk/~vgg/data/vgg_face2/}}.
\end{abstract}


%
\IEEEpeerreviewmaketitle

\section{Introduction}
\label{sec:intro}
\input{texts/introduction.tex}

\section{Dataset Review}
\label{sec:related-work}
\input{texts/dataset_stats.tex}	
\input{texts/related_works.tex}   

\section{An Overview of The \NewVGGFace{}}
\label{sec:overview}
\input{texts/dataset_overview.tex}  

\section{DATASET COLLECTION}
\label{sec:coll-proc}
\input{texts/dataset_collection.tex}  

\section{EXPERIMENTS}
\label{sec:exp}
\input{texts/experiment.tex}	

\section{Conclusion}
\label{sec:conclusion}
\input{texts/conclusion.tex}


\small
\section*{Acknowledgment}
We would like to thank Elancer and 
Momenta for their part in preparing  the dataset\footnote{\url{http://elancerits.com/} ~~~~~ \url{https://momenta.ai/}}.
This research is based upon work supported by the Office of the Director of National Intelligence (ODNI), Intelligence Advanced Research Projects Activity (IARPA), via contract number 2014-14071600010. The views and conclusions contained herein are those of the authors and should not be interpreted as necessarily representing the official policies or endorsements, either expressed or implied, of ODNI, IARPA, or the U.S.\  Government.  The U.S.\  Government is authorized to reproduce and distribute reprints for Governmental purpose notwithstanding any copyright annotation thereon.



%
{\small
\bibliographystyle{ieee}
\bibliography{shortstrings,vgg_local,vgg_other} 
}

\end{document}

%% file: texts/introduction.tex
Concurrent  with the rapid development of deep \CNNs{} (CNNs), 
there has been much recent effort in collecting large scale datasets to feed these data-hungry models.
In general, recent datasets (see Table~\ref{dataset-comp})
have explored the importance of intra- and inter-class variations.
The former focuses on depth (many images of one subject) 
and the latter on breadth (many subjects with limited images per subject).
However, none of these datasets was specifically designed to explore pose and age variation. 
We address that here by designing a dataset generation pipeline to explicitly collect
images with a wide range of pose, age, illumination and ethnicity variations of human faces.

We make the following \emph{four} contributions: 
first, we have collected a new large scale dataset,  \NewVGGFace{}, for public release. 
It includes over nine thousand identities with between 80 and 800 images for each identity, 
and more than 3M images in total; 
second, a dataset generation pipeline is proposed that encourages pose and age diversity for each subject, 
and also involves multiple stages of automatic and manual filtering in order to minimise label noise;  
third, we provide template annotation for the test set to explicitly explore pose and age recognition performance; 
and, finally, we show that training deep CNNs on the new dataset substantially exceeds the state-of-the-art 
performance on the IJB benchmark datasets~\cite{klare2015pushing, whitelam2017iarpa, Maze2018}. 
In particular, we experiment with the recent Squeeze and Excitation network~\cite{jie2017}, 
and also investigate the benefits of first pre-training on a dataset with breadth (MS-Celeb-1M~\cite{guo2016ms})
and then fine tuning on \NewVGGFace{}.

The rest of the paper is organised as follows:
We review previous dataset in Section~\ref{sec:related-work}, 
and give a summary of existing public dataset in Table~\ref{dataset-comp}.
Section~\ref{sec:overview} gives an overview of the new dataset, 
and describes the template annotation for recognition over pose and age.
Section~\ref{sec:coll-proc} describes the dataset collection process. 
Section~\ref{sec:exp} reports state-of-the-art performance of several different architectures on 
the IJB-A~\cite{klare2015pushing}, IJB-B~\cite{whitelam2017iarpa} and IJB-C~\cite{Maze2018} benchmarks.

%% file: texts/dataset_stats.tex
\newcommand{\captiontitle}{Statistics for recent public face datasets}
%
\begin{table*}[ht]
\captionsetup{font=small}
\begin{center}{\scalebox{0.8}{
\begin{tabular}{|l|c|c|c|c|c|c|c|}
\hline
Datasets & \# of subjects & \# of images & \# of images per subject & manual identity labelling & pose & age & year\\ \hline
\hline
LFW~\cite{huang2007labeled} & $5,749$ & $13,233$ & $1$/$2.3$/$530$ & - & - & - & 2007\\ \hline
YTF~\cite{wolf2011face} & $1,595$ & $3,425$ videos & - & - & -  & - & 2011\\ \hline
CelebFaces+~\cite{sun2014deep} & $10,177$ & $202,599$ & $19.9$ & - & - & - & 2014\\ \hline
CASIA-WebFace~\cite{yi2014learning} & $10,575$ & $494,414$ & $2$/$46.8$/$804$ & - & - & - & 2014\\ \hline
IJB-A~\cite{klare2015pushing} & $500$ & $5,712$ images, $2,085$ videos & $11.4$ & - & -& - & 2015\\ \hline
IJB-B~\cite{whitelam2017iarpa} & $1,845$ & $11,754$ images, $7,011$ videos & $36.2$ & - & -& - & 2017\\ \hline
IJB-C~\cite{Maze2018} & $3,531$ & $31,334$ images, $11,779$ videos & $36.3$ & - & - & - & 2018 \\ \hline
VGGFace~\cite{Parkhi15} & $2,622$ & $2.6$ M & $1,000$/$1,000$/$1,000$ & - & -& Yes & 2015\\ \hline
MegaFace~\cite{kemelmacher2016megaface} & $690,572$ & $4.7$ M & $3$/$7$/$2469$ & - & -& - & 2016\\ \hline
MS-Celeb-1M~\cite{guo2016ms} & $100,000$ & $10$ M & $100$ & - & -& - & 2016\\ \hline
UMDFaces~\cite{bansal2016umdfaces} & $8,501$ & $367,920$ & $43.3$ & Yes & Yes & Yes & 2016\\ \hline
UMDFaces-Videos~\cite{bansal2017s} & $3,107$ & $22,075$ videos & - & - &  - & - & 2017\\ \hline
\NewVGGFace { (\textbf{this paper})} & $9,131$ & $3.31$ M & $80$/$362.6$/$843$ & Yes & Yes & Yes & 2018\\ \hline
\end{tabular}}}
\end{center}
\vspace{-1.5mm}
\caption[\captiontitle]{\captiontitle{}. The three entries in the `per subject' column are the  minimum/average/maximum per subject.}
\label{dataset-comp}
\end{table*}

%% file: texts/related_works.tex
In this section we briefly review the principal ``in the wild''  datasets that have appeared recently. 
In 2007, the Labelled Faces in the Wild (LFW) dataset~\cite{huang2007labeled} was released,
containing $5,749$ identities with $13,000$ images. 

The CelebFaces+ dataset~\cite{sun2014deep} was released in 2014, 
with $202,599$ images of $10,177$ celebrities. 
The CASIA-WebFace dataset~\cite{yi2014learning} released the same year
that has $494,414$ images of $10,575$ people. 
The VGGFace dataset~\cite{Parkhi15} released in 2015 has $2.6$ million images covering $2,622$ people, 
making it amongst the largest publicly available datasets. 
The curated version, where label noise is removed by human annotators, 
has $800,000$ images with approximately $305$ images per identity. 
Both the CASIA-WebFace and VGGFace datasets were released for training purposes only.

MegaFace dataset~\cite{kemelmacher2016megaface}  was released in 2016  to evaluate 
face recognition methods with up to a million distractors in the gallery image set. 
It contains $4.7$ million images of $672,057$ identities as the training set.
However, an average of only  $7$ images per identity makes it restricted in its per identity face variation.
In order to study the effect of  pose and age variations in recognising faces,   
the MegaFace challenge~\cite{kemelmacher2016megaface} uses the subsets of FaceScrub~\cite{ng2014data} containing $4,000$ images from $80$ identities and FG-NET~\cite{panis2014overview} containing $975$ images from $82$ identities for evaluation. 


Microsoft released the  large Ms-Celeb-1M dataset~\cite{guo2016ms} in 2016 
with $10$ million images from $100$k celebrities for training and testing. This is a very useful dataset, and we 
employ  it for pre-training in this paper.
However, it has two limitations: 
(i) while it has the largest number of training images, 
the intra-identity variation is somewhat restricted due to an average of $81$ images per person; 
(ii) images in the training set were directly retrieved from a search engine without manual filtering, and
consequently there is  label noise. 
The IARPA Janus Benchmark-A (IJB-A)~\cite{klare2015pushing}, 
Benchmark-B (IJB-B)~\cite{whitelam2017iarpa} and Benchmark-C (IJB-C)~\cite{Maze2018} datasets were released as evaluation benchmarks~(only test) for face detection, recognition and clustering in images and videos. 

Unlike the above datasets which are geared towards image-based face recognition, 
the Youtube Face (YTF)~\cite{wolf2011face} and UMDFaces-Videos~\cite{bansal2017s} datasets aim to recognise faces in unconstrained videos. 
YTF contains $1,595$ identities and $3,425$ videos, 
whilst UMDFaces-Videos is larger with $3,107$ identities and $22,075$ videos (the identities are
a subset of those in UMDFaces~\cite{bansal2016umdfaces}).

Apart form these public datasets,
Facebook and Google have large  in-house datasets.
For instance, 
Facebook~\cite{taigman2015web} trained a face identification model using $500$ million images of over $10$ million subjects. 
The face recognition model by Google~\cite{schroff2015facenet} was trained using $200$ million images of $8$ million identities. 


%% file: texts/dataset_overview.tex
\subsection{Dataset Statistics}
\begin{figure*}[ht!]
\captionsetup{font=small}
\begin{center}
\subfigure[pose statistics]{\includegraphics[width=55mm, height = 40mm]{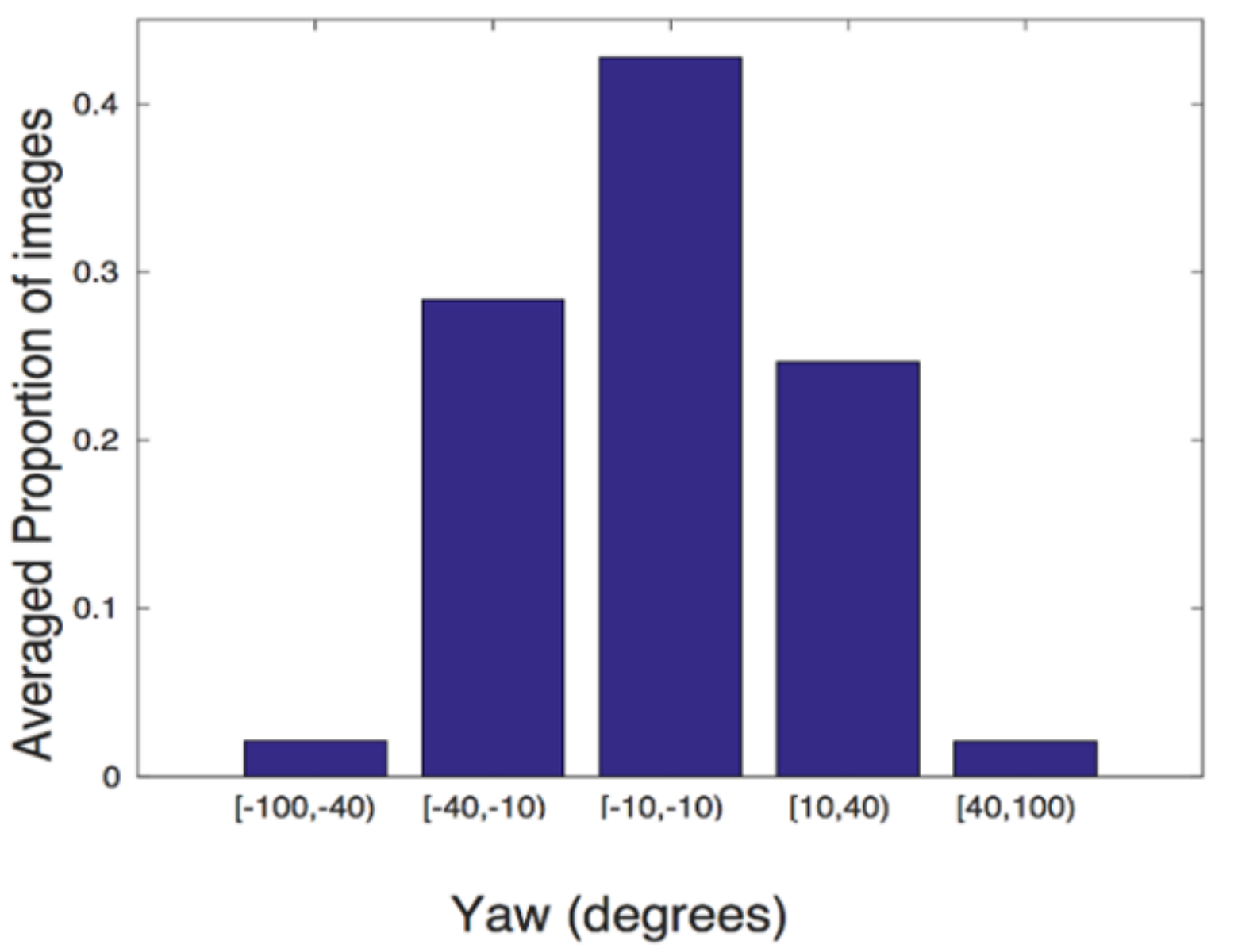}} \hspace{20mm} 
\subfigure[age statistics]{\includegraphics[width=55mm, height = 40mm]{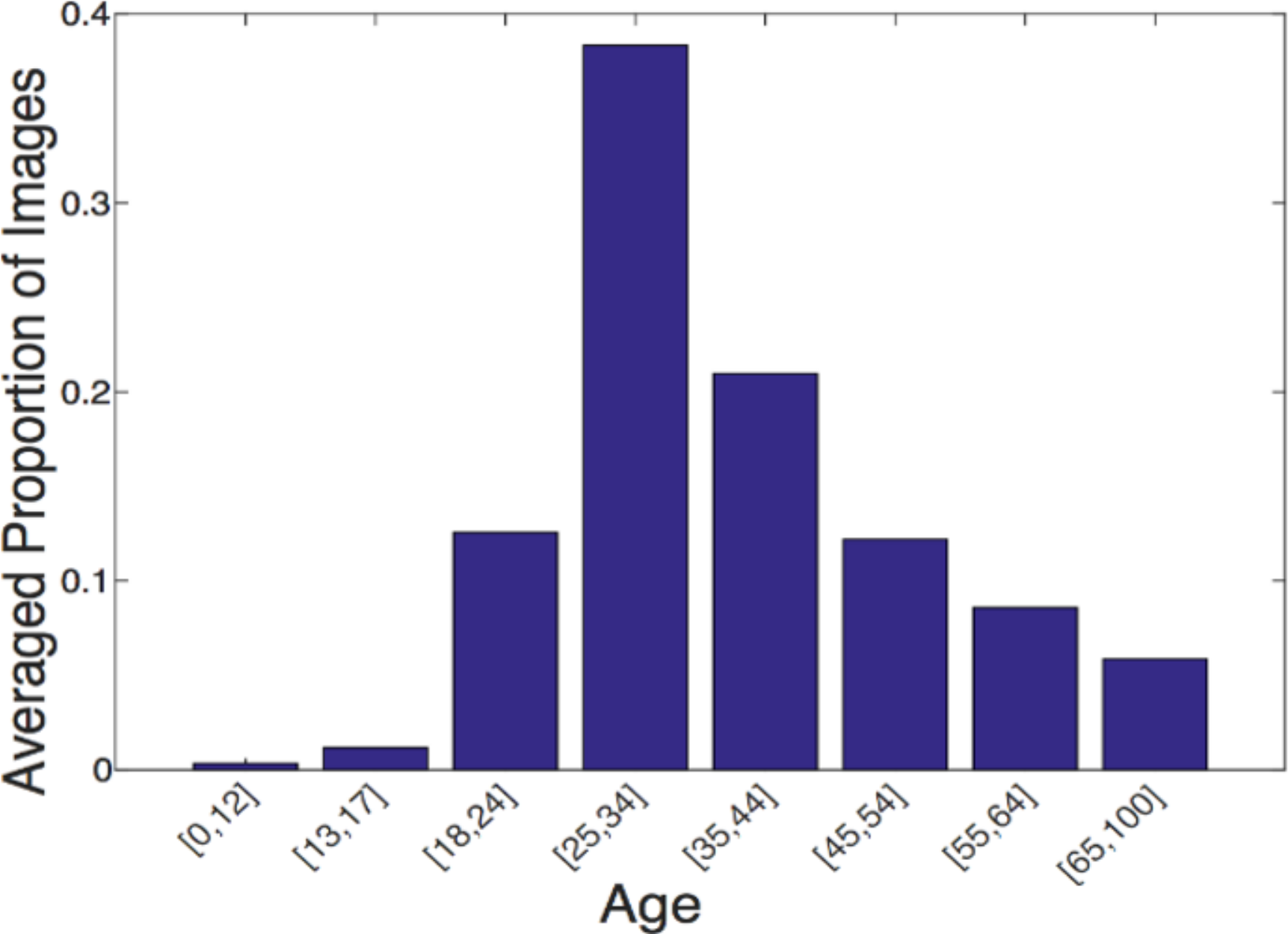}}\\
\subfigure[John Wesley Shipp]{\includegraphics[height=12mm]{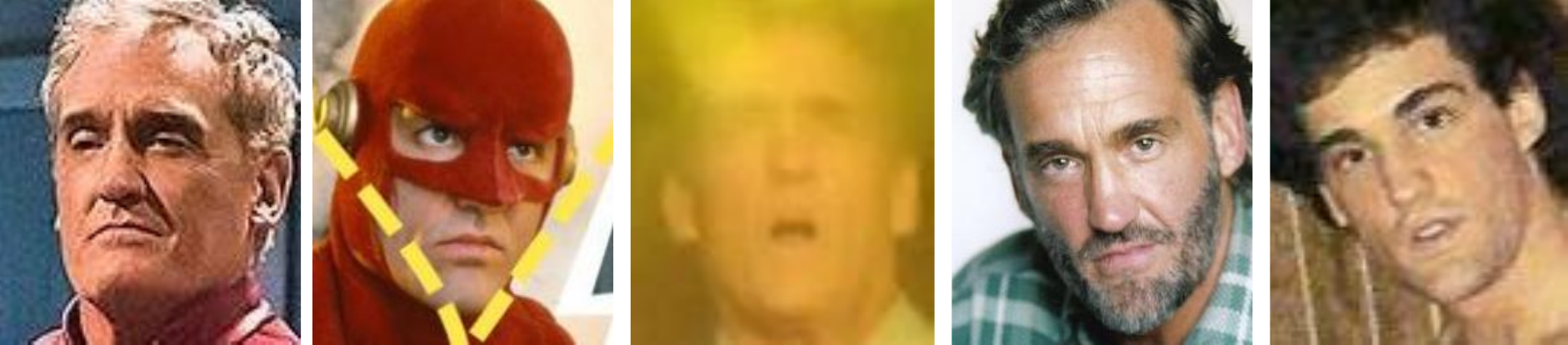}}  \hspace{20mm} 
\subfigure[Leymah Gbowee]{\includegraphics[height=12mm]{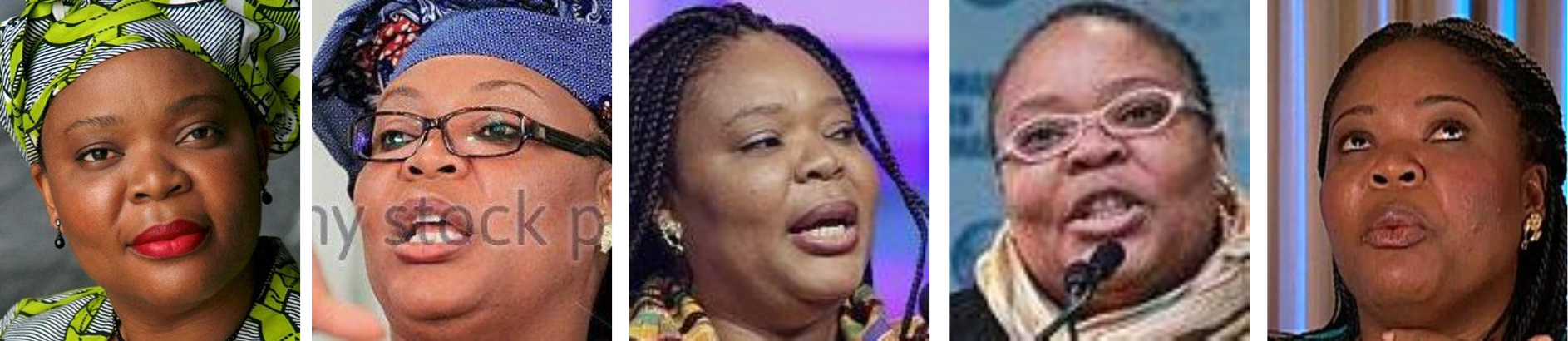}} \\
\subfigure[Princess Haya Bint Al Hussein]{\includegraphics[height=12mm]{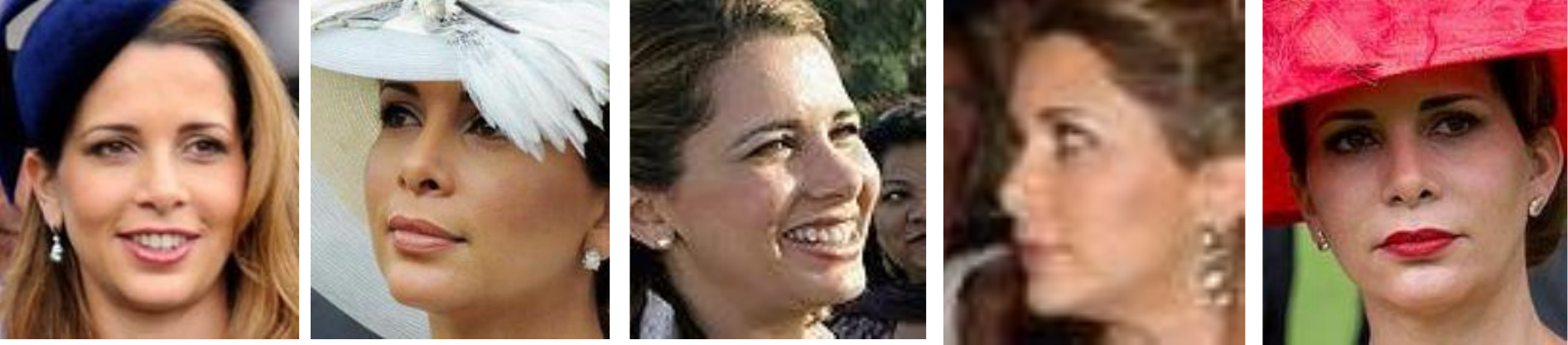}}  \hspace{20mm} 
\subfigure[Julio C\'{e}sar Ch\'{a}vez Jr.]{\includegraphics[height=12mm]{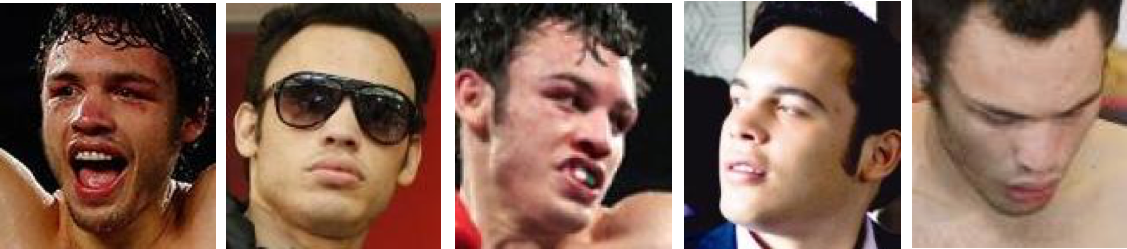}}\\
\subfigure[Roy Jones Jr.]{\includegraphics[height=12mm]{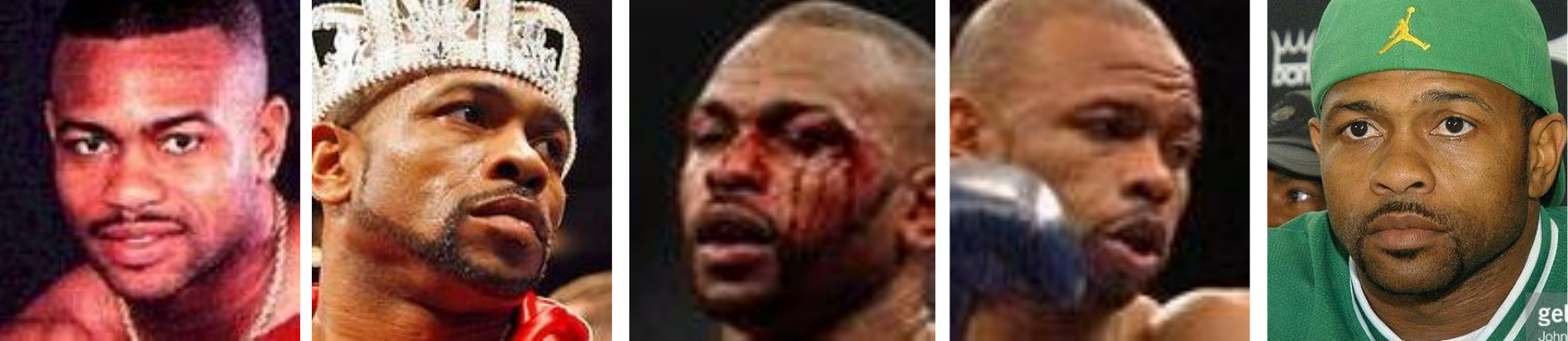}}  \hspace{20mm} 
\subfigure[Ruby Lin]{\includegraphics[height=12mm]{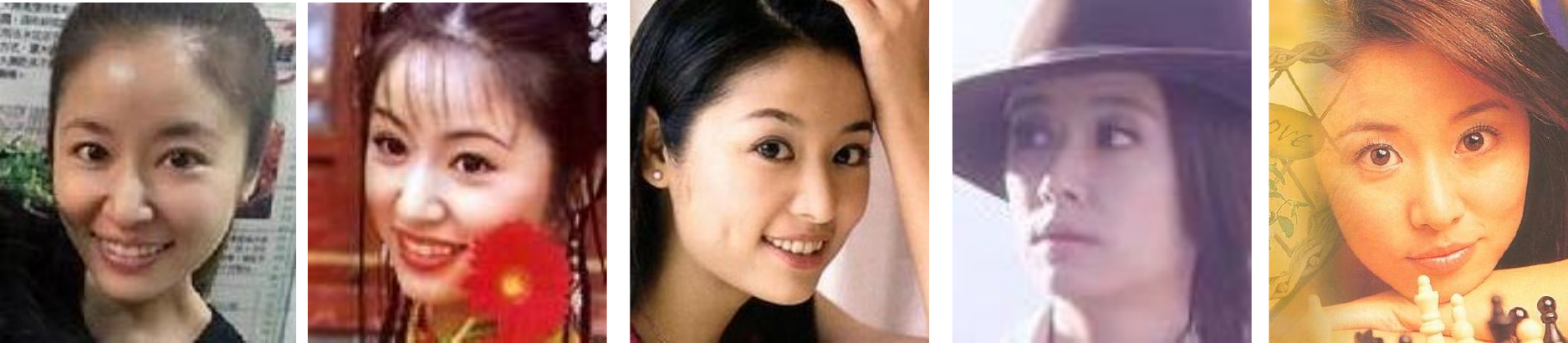}}\\
\subfigure[Additi Gupta]{\includegraphics[height=12mm]{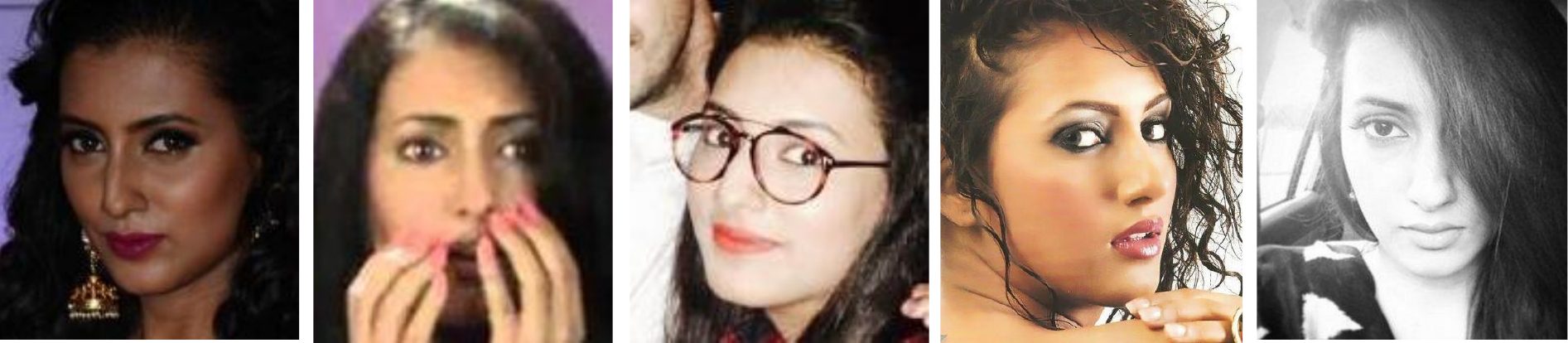}}  \hspace{20mm} 
\subfigure[Lee Joon-gi]{\includegraphics[height=12mm]{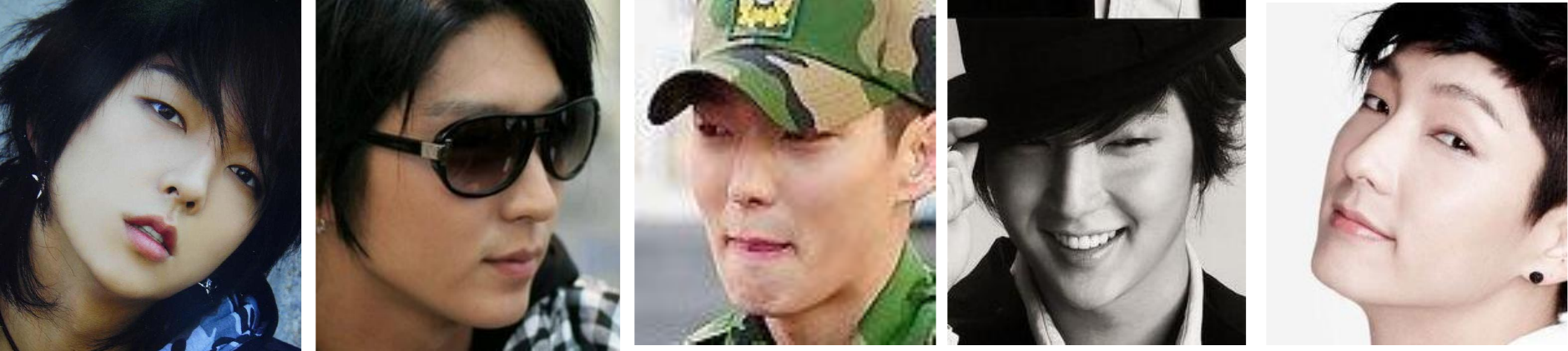}}
\end{center}
\caption{(a-b) \NewVGGFace{} poses and ages statistics. 
(c-j) example images for eight subjects with different ethnicities.}
\label{Random-samples}
\end{figure*}

The \NewVGGFace{} dataset contains $3.31$ million images from $9131$ celebrities
spanning a wide range of ethnicities, e.g.\ it  includes more Chinese and Indian faces than VGGFace 
(though, the ethnic balance is still limited by the distribution of celebrities and public figures),
and professions (e.g.\ politicians and athletes). 
The Images were downloaded from Google Image Search and show large variations in pose, 
age, lighting and background. 
The dataset is approximately gender-balanced, with $59.3$\%  males,
varying between $80$ and $843$ images for each identity, with $362.6$ images on average. 
It includes human verified bounding boxes around faces, 
and five  fiducial keypoints predicted by the model of~\cite{zhang2016joint}.
In addition, 
pose (yaw, pitch and roll) and apparent age information are estimated by our pre-trained pose and age classifiers 
(Pose, age statistics and example images are shown in Figure~\ref{Random-samples}).

The dataset is divided into two splits: 
one for training having $8631$ classes, and one for evaluation (test)  with $500$ classes. 
\input{texts/stage_stats.tex}   

\input{texts/template_example.tex}    

\subsection{Pose and Age Annotations}
\label{sec:annotations}
The \NewVGGFace{} provides annotation to enable evaluation on two scenarios: 
face matching across different poses, and face matching across different ages. \\
\vspace{-0.3mm}
\noindent {\bf Pose templates.}
A template here consists of five faces from the same subject with a consistent pose. 
This pose can be frontal, three-quarter or profile view. 
For a subset of 300 subjects of the evaluation set, 
two templates ($5$ images per template) are provided for each pose view. 
Consequently there are $1.8$K templates with $9$K images in total.
An examples is shown in Figure~\ref{temp-examples} (left). \\
\vspace{-0.3mm}    
\noindent {\bf Age templates.} 
A template here consists of five faces from the same subject with
either an apparent age below 34 (deemed young), or 34 or above (deemed mature).
These are provided for a subset of $100$ subjects from the evaluation set with two templates for each age period,
therefore, there are  $400$ templates with a total of $2$K images. Examples are show in Figure~\ref{temp-examples} (right).

%% file: texts/stage_stats.tex
\renewcommand{\captiontitle}{Dataset statistics after each stage of processing in the collection pipeline}
\begin{table*}[ht]
\captionsetup{font=small}
\begin{center}
\begin{tabular}{|l|l|c|c|r|c|}
\hline
Stage & Aim & Type & \# of subject & total \# of images  & Annotation effort\\ \hline
\hline
1 & Name list selection & M & $500$K & 50.00 million & $3$ months \\ \hline
2 & Image downloading & A & $9244$ & $12.94$ million & - \\ \hline
3 & Face detection  & A & $9244$ & $7.31$ million & -\\ \hline
4 & Automatic filtering by classification  & A & $9244$ & $6.99$ million & -\\ \hline
5 & Near duplicate removal & A & $9244$ & $5.45$ million & - \\ \hline
6 & Final automatic and manual filtering & A/M & $9131$ & $3.31$ million & $21$days\\ \hline
\end{tabular}
\end{center}
\vspace{-1.5mm}
\caption[\captiontitle]{\captiontitle{}.}
\label{stage_statistics}
\end{table*}

%% file: texts/template_example.tex
\renewcommand{\captiontitle}{\NewVGGFace{}  template examples. Left: pose templates from 
three different viewpoints (arranged by row) -- frontal, three-quarter, profile. Right:  age templates for two subjects for young
and mature ages (arranged by row)}
\begin{figure*}[ht]
\captionsetup{font=small}
\begin{center}
\subfigure{\includegraphics[width=70mm]{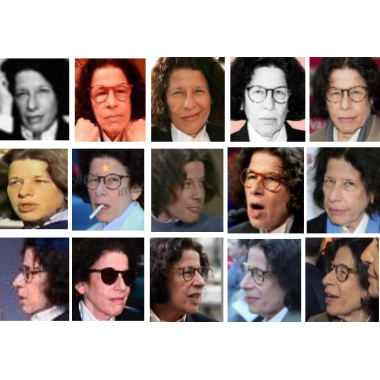}}\hspace{20mm} 
\subfigure{\includegraphics[width=70mm]{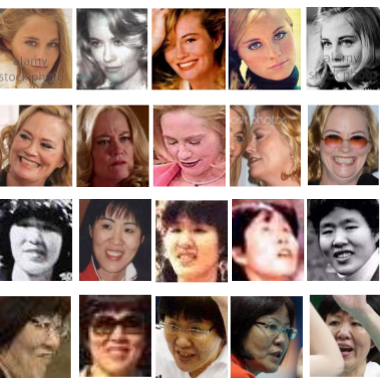}} 
\end{center}
\vspace{-4mm}
\caption[\captiontitle]{\captiontitle{}.}

\label{temp-examples}
\end{figure*}


%% file: texts/dataset_collection.tex
In this section, we describe the dataset collection process, 
including:
how a list of candidate identities was obtained;
how candidate images were collected;
and, how the dataset was cleaned up both automatically and manually.
The process is summarised in Table~\ref{stage_statistics}.

\subsection{Stage 1: Obtaining and selecting a  name  list}
\label{sec:stage1}
We use a similar strategy to that proposed by~\cite{Parkhi15}.
The first stage is to find as many subjects as possible that have a sufficiently distinct
set of images available, for example, celebrities and public figures (e.g. actors, politicians and athletes). 
An initial list of $500$k public figures is obtained from the Freebase knowledge graph~\cite{freebase}. 

An annotator team is then used 
to remove identities from the candidate list that do not have sufficient distinct images. 
To this end, for each of the $500$K names, $100$ images are
downloaded using Google Image Search and human annotators 
are instructed to retain subjects for which 
approximately $90$\% or more of the $100$ images belong to a single identity. 
This removes candidates who do not have sufficient images or for which
Google Image Search returns a mix of people for a single name. 
In this manner, we reduce the candidates to only $9244$ names. 
Attribute information such as ethnicity and kinship is obtained from DBPedia~\cite{dbpedia}.
\subsection{Stage 2: Obtaining images for each identity}
\label{sec:stage2}
We query in Google Image Search and download $1000$ images for each subject. 
To obtain images with large pose and age variations, 
we then append the keyword `sideview' and `very young' to each name and download $200$ images for each. 
This results in $1400$ images for each identity.        

\subsection{Stage 3: Face detection}
\label{sec:stage3}
Faces are detected using the model provided by~\cite{zhang2016joint}.
We use the hyper-parameters recommended in that work to favor a good trade-off between precision and recall. 
The face bounding box is then extended by a factor of $0.3$ to include the whole head. 
Moreover, five facial landmarks are predicted by the same model.

\subsection{Stage 4: Automatic filtering by classification}
\label{sec:stage4}
The aim of this stage is to remove outlier faces for each identity automatically. 
This is achieved by learning a classifier to identify the faces, 
and removing possible erroneous faces below a classification score.
To this end, 1-vs-rest classifiers are trained to discriminate between the $9244$ subjects. 
Specifically, faces from the top $100$ retrieved images of each identity are used as positives, 
and the top $100$ of all other identities are used as negative for training. The
face descriptor features  are obtained from
the VGGFace~\cite{Parkhi15}  model.
Then, the scores (between $0$ and $1$) from the trained model is used to sort images for each subject 
from most likely to least likely. 
By manually checking through images from a random $500$ subjects, 
we choose a threshold of $0.5$ and remove any faces below this.

\subsection{Stage 5: Near duplicate removal}
\label{sec:stage5}
The downloaded images also contain exact or near duplicates due to
the same images being found at different internet locations, 
or images differing only slightly in colour balance or JPEG artifacts for example. 
To alleviate this,  duplicate images are removed  by clustering  VLAD
descriptors for all images remaining at \emph{stage $4$} 
and only retaining one image per cluster~\cite{arandjelovic13,jegou2012aggregating}.

\subsection{Stage 6: Final  automatic and manual filtering}
\label{sec:stage6}
At this point,  
two types of error may still remain: 
first, some classes still have outliers (i.e.\ images that do not belong to the person);
and second, some classes contain a mixture of faces of more than one person, 
or they overlap with another class in the dataset. 
This stage addresses these two types of errors with a mix of manual and automated algorithms.\\

\noindent {\bf Detecting overlapped subjects.} 
Subjects may overlap with other subjects. 
For instance, 
`Will I Am' and `William James Adams' in the candidate list refer to the same person. 
To detect confusions for each class, 
we randomly split the data for each class in half: 
half for training and the other for testing. 
Then, we train a ResNet-50~\cite{he2016deep} and generate a confusion matrix by calculating top-1 error
on the test samples. 
In this manner, we find $20$ subjects confused with others. 
In this stage, we removed $19$ noisy classes. 
In addition, we remove $94$ subjects with samples less than $80$ images, which results in a final list of $9131$ identities.\\

\noindent {\bf Removing outlier images for a subject.} 
The aim of this filtering, which is partly manual,  is to achieve a purity greater than $96$\%. 
We found that for some subjects, images with very high classifier scores at \emph{stage $4$} can also be noisy. 
This happens when the downloaded images contain couples or band members who always appear together in public. 
In this case, the classifiers trained with these mixed examples at \emph{stage $4$} tend to fail.

We retrain the model based on the current dataset,
and for each identity
the classifier score is used to divide the images into $3$ sets: 
H (i.e.\ high score range [$1$, $0.95$]), 
I (i.e.\ intermediate score range ($0.95$, $0.8$]) and 
L (i.e.\ low score range ($0.8$, $0.5$]).
Human annotators  clean up the images for each subject based on their scores,
and the actions they carry out depends on whether the set H is noisy or not. 
If the set (H) contains several different people (noise) in a single identity folder,
then set I and L (which have lower confidence scores), will undoubtedly be noisy as well, 
so all three sets are cleaned manually.
In contrast, if set H is clean, then only set L  (the lowest scores which is supposed to be the most noisy set) is cleaned up.
After this, a new model is trained on the cleaned set H and L,
and set I (intermediate scores,  noise level is also intermediate) is then cleaned by model prediction. This procedure
achieves very low label noise without requiring manual checking of every image.

\subsection{Pose and age annotations}
\label{sec:pose-age-annotations}
We train two networks to obtain the pose and age information for the  dataset. 
To obtain head pose (roll, pitch, yaw), 
a $5$-way classification ResNet-$50$~\cite{he2016deep} is trained 
on the CASIA-WebFace dataset~\cite{yi2014learning}. 
Then, this trained model is used to predict pose for all the images in the dataset.

Similarly, to estimate the apparent age, 
a $8$-way classification ResNet-$50$~\cite{he2016deep} is trained 
on IMDB-WIKI - $500$k$+$ dataset~\cite{rothe2015dex}. 
Ages of faces are then predicted by this model. 

%% file: texts/experiment.tex
In this section, we evaluate the quality of the \NewVGGFace{} dataset by conducting  a number of  baseline experiments. 
We report the results on the \NewVGGFace{} test  set,  
and evaluate on the  public benchmarks IJB datasets\cite{klare2015pushing, whitelam2017iarpa, Maze2018}.  
The subjects in our training dataset are disjoint with the ones in benchmark datasets. 
We also remove the overlap between MS-Celeb-1M and the two benchmarks when training the networks.
\subsection{Experimental setup}
\label{sec:exp-setup}
\noindent {\bf Architecture.} 
ResNet-50~\cite{he2016deep} and SE-ResNet-50~\cite{jie2017} 
(SENet for short) are used as the backbone architectures for the comparison amongst training datasets.  
The Squeeze-and-Excitation (SE) blocks~\cite{jie2017} adaptively recalibrate channel-wise feature responses
by explicitly modelling channel relationships. 
They can be integrated with modern architectures, such as ResNet, 
and improve its representational power. 
This has been demonstrated for object and scene classification, 
with a Squeeze-and-Excitation network winning the ILSVRC 2017 classification competition.

The following experiments are developed under four settings: 
(a) networks are learned from scratch on VGGFace~\cite{Parkhi15} (VF for short); 
(b) networks are learned from scratch on MS-Celeb-1M (MS1M for short)~\cite{guo2016ms}; 
(c) networks are learned from scratch on \NewVGGFace{} (\shortNewVGGFace{} for short); and,
(d) networks are first pre-trained on MS1M, and then fine-tuned on \NewVGGFace{} (\shortNewVGGFace{}\_ft for short).\\

\noindent {\bf Similarity computation.}
In all the experiments~(i.e.\ for both verification and
identification), we need to compute the similarity between subject
templates.  A template
is represented by a single vector computed by aggregating the face
descriptors of each face in the template set.  In
section~\ref{sec:exp-\NewVGGFace{}}, the template vector is obtained
by averaging the face descriptors of the images and SVM classifiers
are used for identification.  In sections~\ref{sec:exp-ijba}
and~\ref{sec:exp-ijbb} for IJB-A and IJB-B,  where the template may
contain both still images and video frames, we first compute the media
vector (i.e.\ from images or video frames) by averaging the face descriptors  in
that media. A template vector is then generated by averaging the media
vectors in that template, which is then L$2$ normalised.
Cosine similarity is used to represent the similarity between two templates.

A face descriptor is obtained from the trained networks as follows: 
first the
extended bounding box of the face is resized so that 
the shorter side is $256$ pixels; then
the centre $224\times 224$ crop of the face image is used as input to the network.
The face descriptor is extracted from from the layer adjacent to the classifier layer. 
This leads to a $2048$ dimensional descriptor, which is then L$2$ normalised.\\

\noindent {\bf Training implementation details.} 
All the networks are trained for classification using the soft-max loss function.
During training, the
extended bounding box of the face is resized so that 
the shorter side is $256$ pixels,  then
a  $224\times 224$ pixels region is randomly
cropped from each sample. The mean
value of each channel is subtracted for each pixel.

Monochrome augmentation is used with a probability of $20\%$ to reduce the over-fitting on colour images. 
Stochastic gradient descent is used with mini-batches of size $256$, 
with a balancing-sampling strategy for each mini-batch due to the unbalanced training distributions.
The initial learning rate is $0.1$ for the models trained from scratch, 
and this is decreased twice with a factor of $10$ when errors plateau. 
The weights of the models are initialised as described in~\cite{he2016deep}. 
The learning rate for model fine-tuning starts from $0.005$ and decreases to $0.001$.

\subsection{Experiments on the new dataset}
\label{sec:exp-\NewVGGFace{}}
In this section, we evaluate ResNet-50 trained from scratch on the three datasets
as described in the Sec.~\ref{sec:exp-setup},
and~\NewVGGFace{} test set.
We test identification performance and also similarity over pose and age, 
and validate the capability of \NewVGGFace{} to tackle  pose and age variations. \\

\noindent {\bf Face identification.}
This scenario aims to predict, for a given test image, whose face it is. 
Specifically, for each of the $500$ subjects in the evaluation set, 
$50$ images are randomly chosen as the testing split and the remaining images are used as the training split. 
This training split is used to learn 1-vs-rest SVM classifiers for each subject.
A top-1 classification error is then used to evaluate the performance of these classifiers on the test images.
As shown in Table~\ref{faceidentification-results},  
there is a significant improvement for the model trained on \NewVGGFace{} rather than on VGGFace. 
This demonstrates the benefit of increasing data variation (e.g, subject number, pose and age variations) in 
the \NewVGGFace{} training dataset. 
More importantly, models trained on \NewVGGFace{} also achieve better result than that on MS1M even though it 
has tenfold more subjects and threefold more images, demonstrating the good quality of \NewVGGFace.
In particular, the very low top-1 error of \NewVGGFace{} provides evidence that there is very little label noise
in the dataset -- which is one of our design goals.
\input{texts/face_identification_vgg2.tex}\\

\input{texts/probing_pose.tex}
\noindent {\bf Probing across pose.} 
This test aims to assess how well templates match across three pose views: front, three-quarter and profile views.    
As described in section~\ref{sec:annotations}, 
$300$ subjects in the evaluation set are annotated with pose templates, 
and there are six templates for each subject: 
two each for front, three-quarter view and profile views. 

These six templates are divided into two sets, one pose for each set,
and a $3 \times 3$ similarity matrix is constructed between the two sets. 
Figure~\ref{pose-temp-analysis-samples} visualises two example of these cosine similarity 
scores for front-to-profile templates.

Table~\ref{tab:faceprobe-pose-result}  compares the similarity matrix averaged over the $300$ subjects.
We can observe that 
(i) all the three models perform better when matching similar poses, 
i.e., front-to-front, three-quarter-to-three-quarter and profile-to-profile; and
(ii) the performance drops when probing for different poses, e.g., front-to-three-quarter and front-to-profile,
showing that recognition across poses is a much harder problem. 
Figure~\ref{pose-analysis} shows histograms of similarity scores. 
It is evident that the mass of the \NewVGGFace{} trained model is to the right of the MS1M and VGGFace trained models. 
This clearly demonstrates the benefit of training on a dataset with larger pose variation.\\

\input{texts/probing_age.tex}  		
\noindent {\bf Probing across age.} 
This test aims to assess how well templates match across age, for two ages ranges: young and mature ages.
As described in section~\ref{sec:annotations}, 
$100$ subjects in the evaluation set are annotated with age templates, 
and there are four templates for each subject: two each for young and mature faces.

For each subject a $2 \times 2$ similarity matrix is computed, 
where an element is the cosine similarity between two templates.  
Figure~\ref{age-temp-analysis-samples} shows two examples of the young-to-mature templates, and their similarity scores.

Table~\ref{tab:faceprobe-age-result} compares the similarity matrix averaged over the $100$ subjects as the model changes.  
For all the three models, there is always a big drop in performance when matching across young and mature faces, 
which reveals that young-to-mature matching is substantially more challenging than young-to-young and mature-to-mature.  
Moreover,
young-to-young matching is more difficult than mature-to-mature matching. 
Figure~\ref{age-analysis} illustrates the histograms of the young-to-mature template similarity scores.\\

\noindent {\bf Discussion.} 
In the evaluation of pose and age protocols,
models trained on \NewVGGFace{} always achieve the highest similarity scores,
and MS1M dataset the lowest. 
This can be explained by the fact that the MS1M dataset is designed to focus more on inter-class diversities,
and this harms the matching performance across different pose and age,
illustrating the value of \NewVGGFace{} in having more intra-class diversities that cover large variations in pose and age.  

\input{texts/template_matches.tex}

\subsection{Experiments on IJB-A}
\label{sec:exp-ijba}
\input{texts/IJBA_result.tex}  
In this section, we compare the performance of the models trained on the different datasets on the public IARPA Janus Benchmark A (IJB-A dataset)~\cite{klare2015pushing}. 

The IJB-A dataset contains $5712$ images and $2085$ videos from $500$ subjects, 
with an average of 11.4 images and 4.2 videos per subject. 
All images and videos are captured from unconstrained environment and show large variations in expression and image qualities. 
As a pre-processing, 
we detect the faces using MTCNN~\cite{zhang2016joint} to keep the cropping consistent between training and evaluation. 

IJB-A provides ten-split evaluations with two standard protocols, namely, 1:1 face verification and 1:N face identification, 
where we directly extract the features from the models for the test sets and use cosine similarity score.
For verification, the performance is reported using the true accept rates (TAR) vs.\  false positive rates (FAR) (i.e.\  receiver operating characteristics (ROC) curve).
For identification, the performance is reported using the true positive identification rate (TPIR) vs.\  false positive identification rate (FPIR) (equivalent to a decision error trade-off (DET) curve) and the Rank-N (i.e. the cumulative match characteristic (CMC) curve).
Table~\ref{jiba-evaluation-table} and Figure~\ref{ijba-curves} presents the comparison results. \\

\noindent{\bf The effect of training set.} 
We first investigate the effect of different training sets based on the same architecture ResNet-50 (Table~\ref{jiba-evaluation-table}), and start with networks trained from scratch.
we can observe that the model trained on \NewVGGFace{} outperforms the one trained on VGGFace by a large 
margin, even though VGGFace has a similar scale (2.6M images) it has fewer identities and pose/age 
variations (and more label noise).
Moreover, 
the model of \NewVGGFace{} is significantly superior to the one of MS1M which has $10$ times subjects over our dataset. 
Specially, it achieve $\sim 4.4\%$ improvement over MS1M on FAR=$0.001$ for verification, 
$\sim 3.7\%$ on FPIR=$0.01$ and $\sim 1.5\%$ on Rank-$1$ for identification. 

When comparing with the results of existing works,
the model trained on \NewVGGFace{} surpasses previously reported results on all metrics 
(best to our knowledge, reported on IJB-A 1:1 verification and 1:N identification protocols), 
which further demonstrate the advantage of the \NewVGGFace{}  dataset.
In addition, 
the generalisation power can be further improved by first training with MS1M and then fine-tuning with \NewVGGFace{}
(i.e.\  ``\NewVGGFace{}\_ft"),
however, the difference is only $0.908$ vs.\  $0.895$.

Many existing datasets are constructed by following the assumption of the superiority of wider dataset (more identities)~\cite{yi2014learning, guo2016ms, kemelmacher2016megaface}, 
where the huge number of subjects would increase the difficulty of  model training. 
In contrast, \NewVGGFace{} takes  both aspects of breath (subject number) and depth (sample number per subject) into account, guaranteeing rich intra-variation and inter-diversity.\\

\noindent{\bf The effect of architectures.} 
We next  investigate the effect of architectures trained on \NewVGGFace{} (Table~\ref{jiba-evaluation-table}). 
The comparison between ResNet-50 and SENet both learned from scratch reveals that SENet 
has a consistently superior performance on both verification and identification.  
More importantly, SENet trained from scratch achieves comparable results to the fine-turned ResNet-50 (i.e.\ first pre-trained on the MS1M dataset),
demonstrating that the diversity of our dataset can be further exploited by an advanced network.  
In addition, the performance of SENet can be further improved by training on the two datasets
\NewVGGFace{} and MS1M, exploiting the different advantages that each offer.

\subsection{Experiments on IJB-B}
\label{sec:exp-ijbb}
\input{texts/IJBB_result.tex}
The IJB-B dataset is an extension of IJB-A, having $1,845$ subjects with $21.8$K still images (including $11, 754$ face and $10,044$ non-face) and $55$K frames from $7,011$ videos.  
We evaluate the models on the standard 1:1 verification protocol (matching between the Mixed Media probes and two galleries) 
and 1:N identification protocol (1:N Mixed Media probes across two galleries).

We observe a similar behaviour to that of the IJB-A evaluation.
For the comparison between different training sets (Table~\ref{jibb-evaluation-table} and Figure~\ref{ijbb-curves}), 
the models trained on \NewVGGFace{} significantly surpass the ones trained on MS1M, 
and the performance can be further improved by integrating the advantages of the two datasets.  
In addition, SENet's superiority over ResNet-50 is evident in both verification and identification with the
two training settings (i.e.\  trained from scratch and
fine-tuned). 
Moreover, we also compare to the results reported by others on the benchmark~\cite{whitelam2017iarpa} (as shown in
Table~\ref{jibb-evaluation-table}), 
and there is a considerable improvement over their performance for all  measures.

\subsection{Experiments on IJB-C}
\label{sec:exp-ijbc}
\input{texts/IJBC_result.tex}
The IJBC dataset is a further extension of IJB-B, including $3531$ subjects with $31.3$K still images and $117.5$K frames from $11,779$ videos. We evaluate the models on the standard 1:1 verification protocol and 1:N identification protocol. Results are shown in Table~\ref{jibc-evaluation-table} and Figure~\ref{ijbc-curves}. Compared to the results reported in ~\cite{Maze2018}, there is a considerable improvement for all measures.

%% file: texts/face_identification_vgg2.tex
\begin{table}[ht!]
\captionsetup{font=small}
\begin{center}
\begin{tabular}{|l|c|c|c|c|}
\hline
 Training dataset & VGGFace & MS1M & \NewVGGFace \\ \hline
\hline
Top-1 error (\%) & $10.6$ & $5.6$ & $3.9$ \\ \hline
\end{tabular}
\end{center}
\caption{Identification performance (top-1 classification error)  on the \NewVGGFace{} test set for ResNet models trained on different datasets.  A lower value is better.}
\label{faceidentification-results}
\end{table}

%% file: texts/probing_pose.tex
\renewcommand{\captiontitle}{Face probing across poses}
\begin{table*}[ht!]
\captionsetup{font=small}
\begin{center}{\scalebox{0.76}{
\begin{tabular}{|c|c|c|c|c|c|c|c|c|c|}
\hline
Training dataset & \multicolumn{3}{c|}{VGGFace}  &  \multicolumn{3}{c|}{MS1M} &  \multicolumn{3}{c|}{\NewVGGFace} \\
 & front & three-quarter & profile
& front & three-quarter & profile
& front & three-quarter & profile \\
\hline
front &  $0.5781$ & $0.5679$& $0.4821$ &$0.5661$ & $0.5582$& $0.4715$ &  $0.6876$ & $0.6821$& $0.6222$ \\
three-quarter &  $0.5706$ & $0.5957$& $0.5345$ & $0.5628$ & $0.5766$& $0.5036$ & $0.6859$ & $0.6980$& $0.6481$\\
profile  &  $0.4859$ & $0.5379$& $0.5682$ & $0.4776$ & $0.5064$& $0.5094$ & $0.6264$ & $0.6515$& $0.6488$ \\
 \hline
\end{tabular}}}
\end{center}
\vspace{-2mm}
\caption[\captiontitle]{\captiontitle{}. Similarity scores are evaluated across pose templates. A higher value is better.}
\label{tab:faceprobe-pose-result}
\end{table*}

\renewcommand{\captiontitle}{Histograms of similarity scores for front-to-profile matching}
\begin{figure*}[ht!]
\captionsetup{font=small}
\begin{center}
\subfigure{\includegraphics[width=50mm]{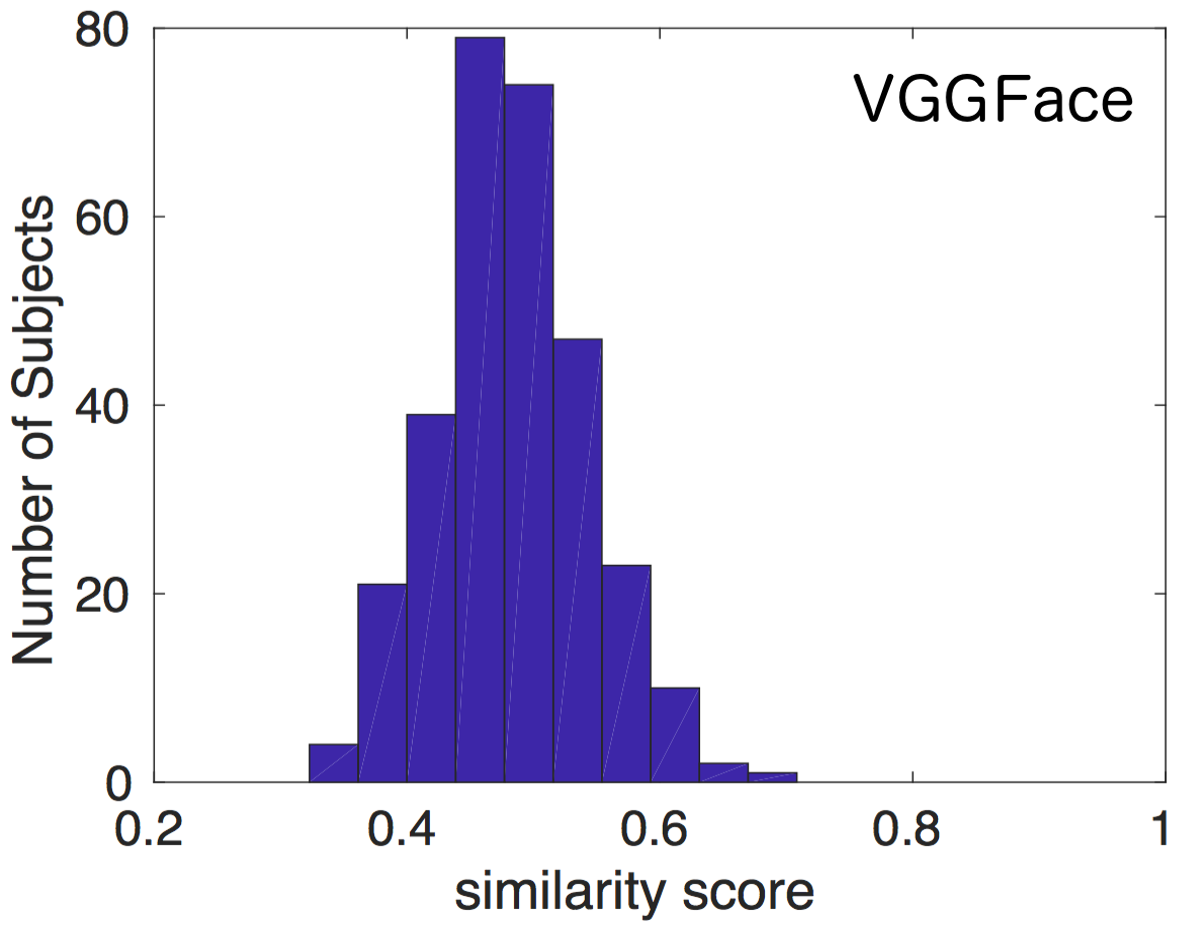}}  
\subfigure{\includegraphics[width=50mm]{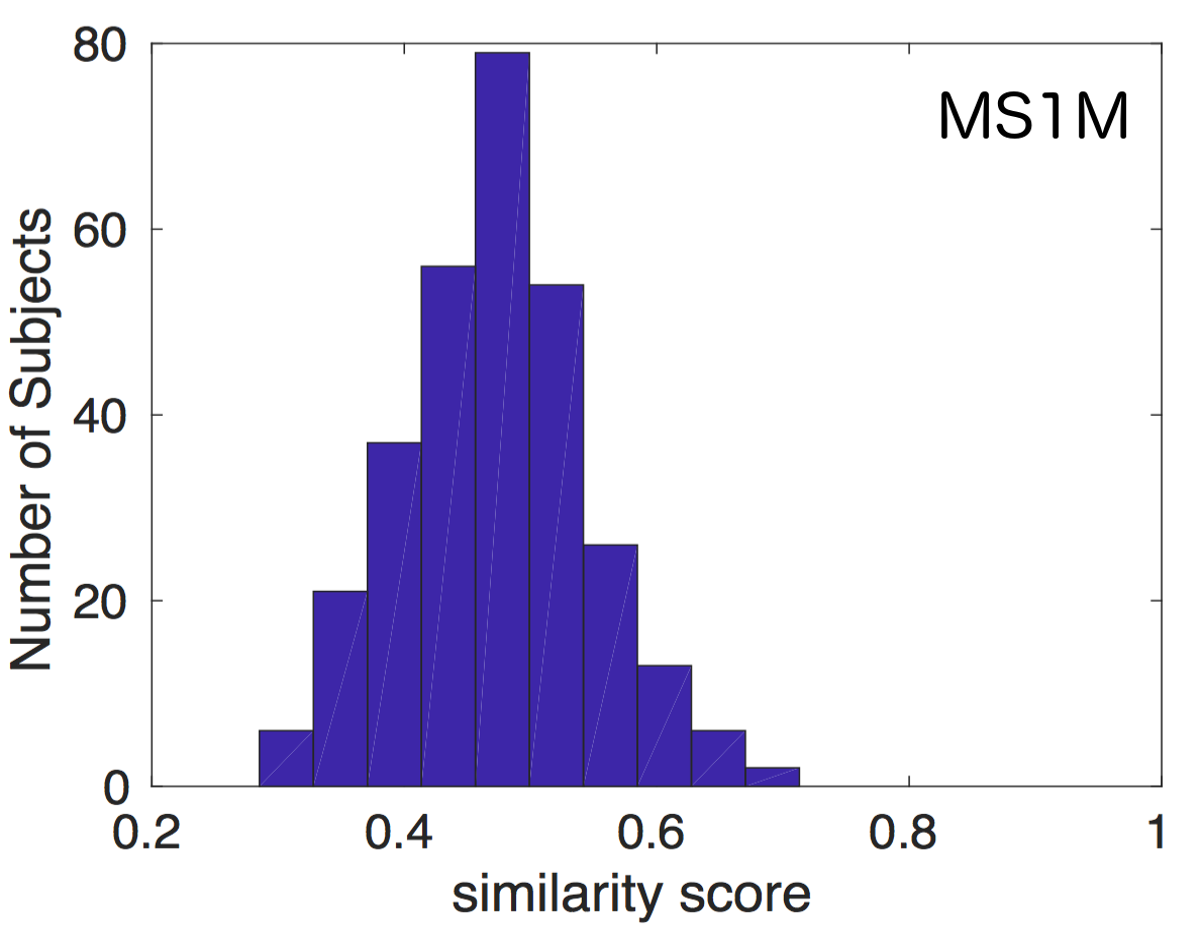}} 
\subfigure{\includegraphics[width=50mm]{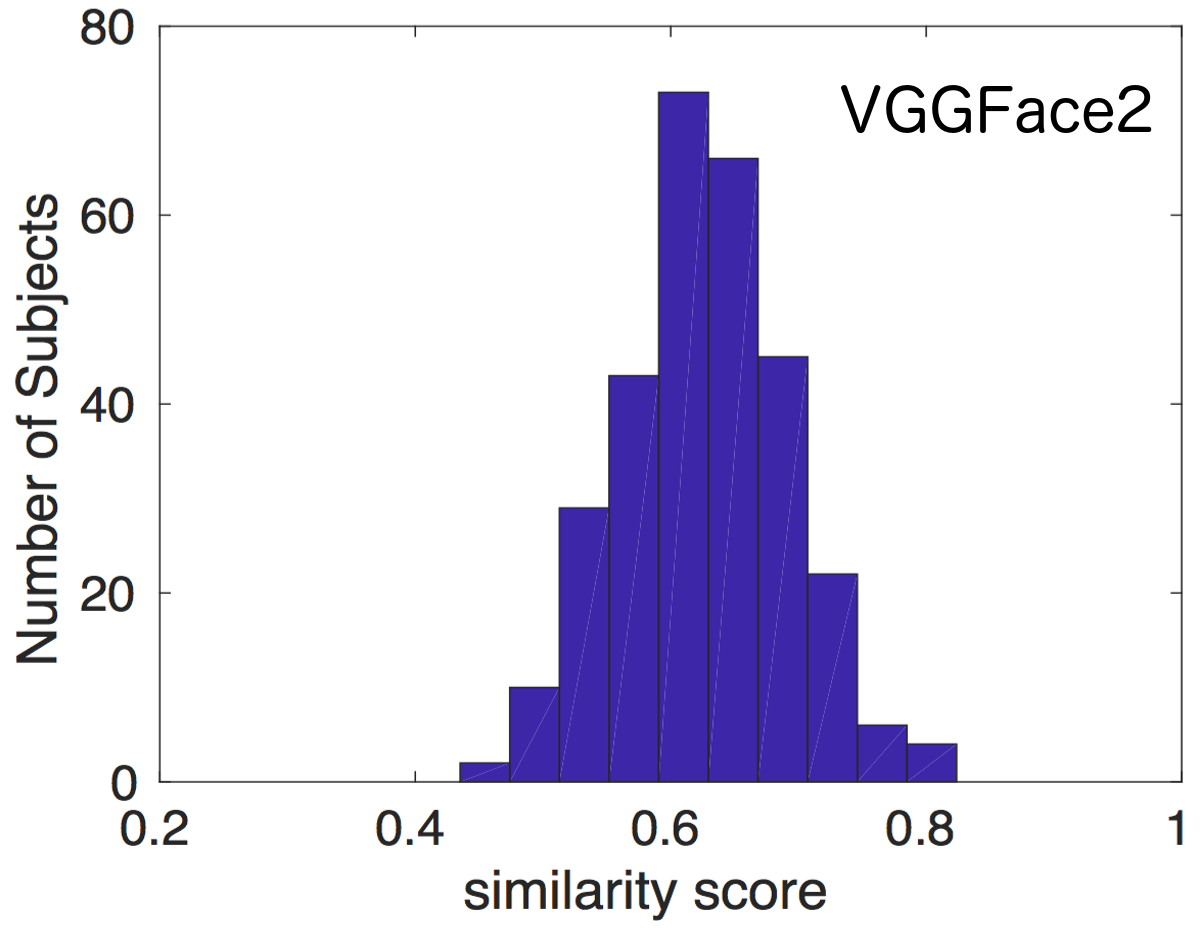}}
\end{center}
\vspace{-3mm}
\caption[\captiontitle]{\captiontitle{} for the models trained on different datasets.}
\label{pose-analysis}
\end{figure*}

\renewcommand{\captiontitle}{Two example templates of front-to-profile matching}
\begin{figure*}[ht!]
\captionsetup{font=footnotesize}
\begin{center}
\subfigure{\includegraphics[width=50mm]{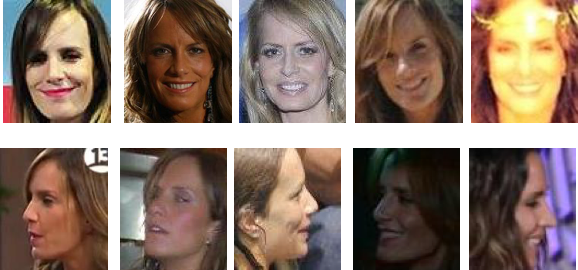}}  \hspace{20mm}
\subfigure{\includegraphics[width=50mm]{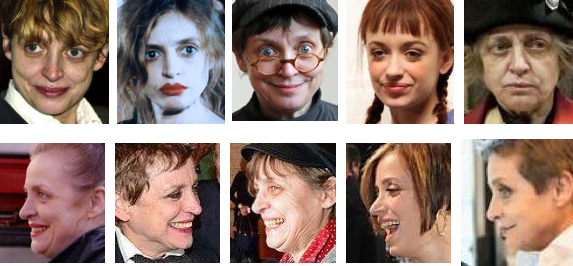}} 
\end{center}
\vspace{-3mm}
\caption[\captiontitle]{\captiontitle{}. Left: the similarity scores produced by VGGFace, MS1M, \NewVGGFace{} are $0.41$, $0.35$ and $0.59$, respectively; Right:  the scores are $0.41$, $0.31$ and $0.57$, respectively.}
\label{pose-temp-analysis-samples}
\end{figure*}

%% file: texts/probing_age.tex
\begin{table*}[ht!]
\captionsetup{font=small}
\begin{center}{\scalebox{0.9}{
\begin{tabular}{|c|c|c|c|c|c|c|c|c|c|}
\hline
 Training dataset & \multicolumn{2}{c|}{VGGFace} &  \multicolumn{2}{c|}{MS1M} &  \multicolumn{2}{c|}{\NewVGGFace} \\
 & young & mature
& young & mature 
& young & mature \\
\hline
young &  $0.5231$ & $0.4338$ & $0.4983$& $0.4005$ & $0.6256$ &  $0.5524$\\
mature &  $0.4394$ & $0.5518$& $0.4099$& $0.5276$ & $0.5607$ &  $0.6637$\\
 \hline
\end{tabular}}}
\end{center}
\vspace{-2mm}
\caption{Face probing across ages. Similarity scores are evaluated across age templates. A higher value is better.}
\label{tab:faceprobe-age-result}
\end{table*}

\begin{figure*}[ht!]
\captionsetup{font=footnotesize}
\begin{center}
\subfigure{\includegraphics[width=50mm]{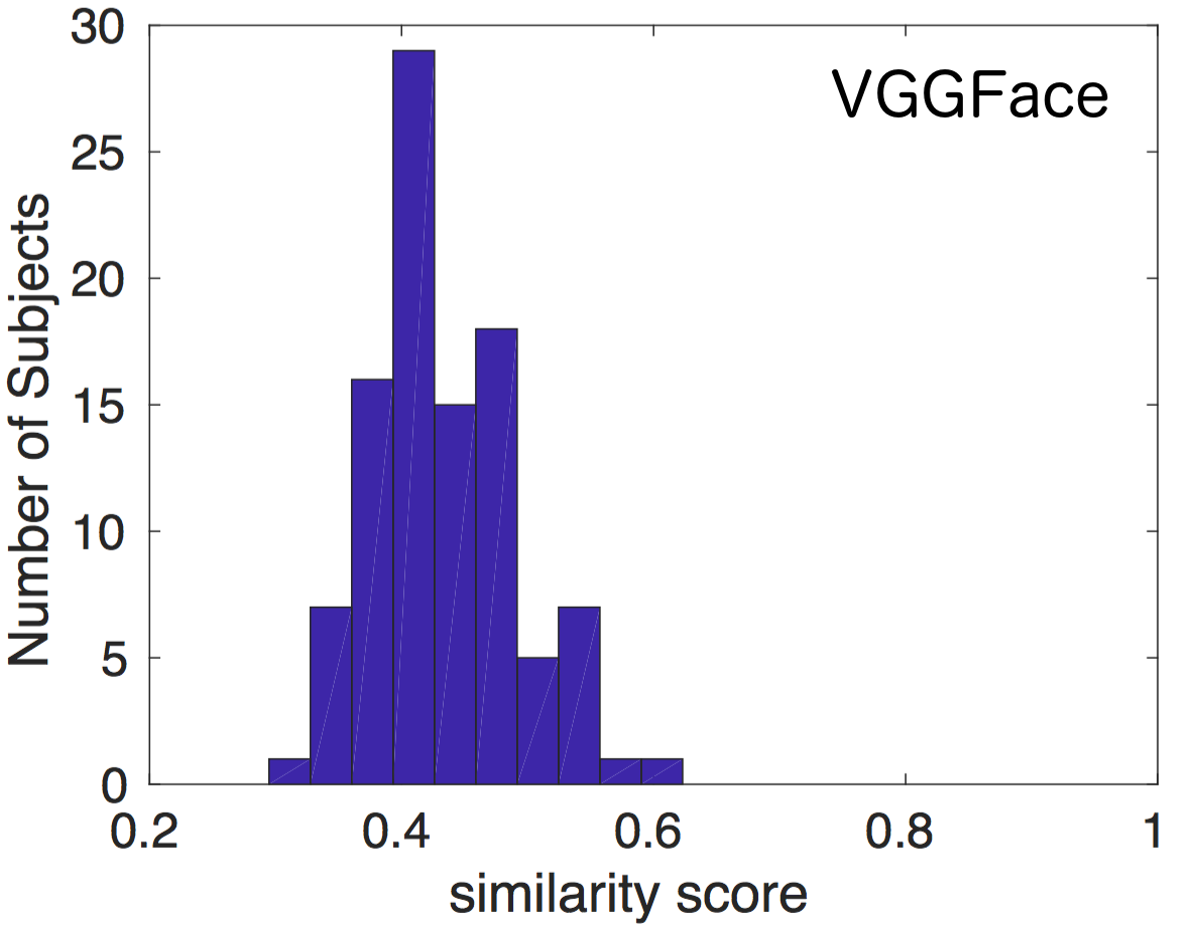}}  
\subfigure{\includegraphics[width=50mm]{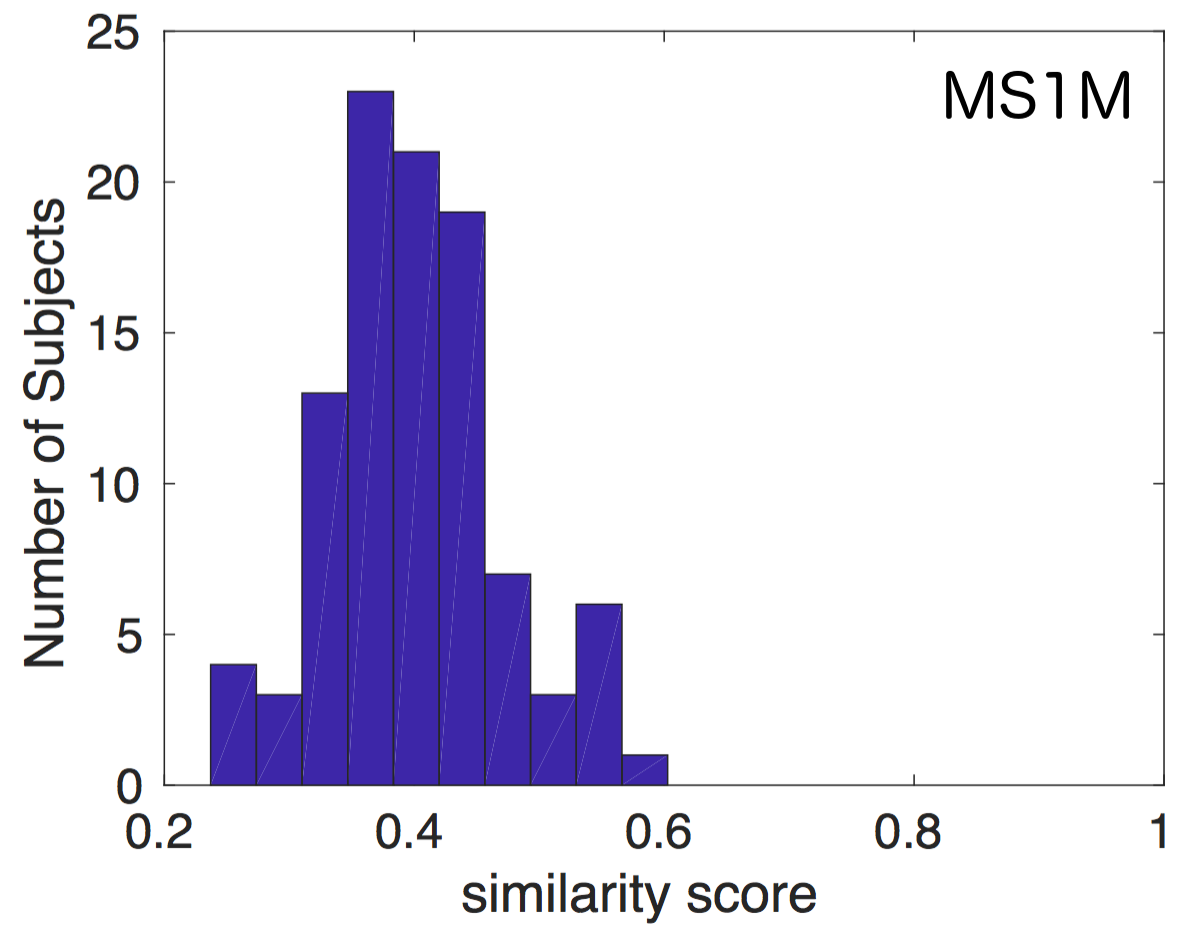}} 
\subfigure{\includegraphics[width=50mm]{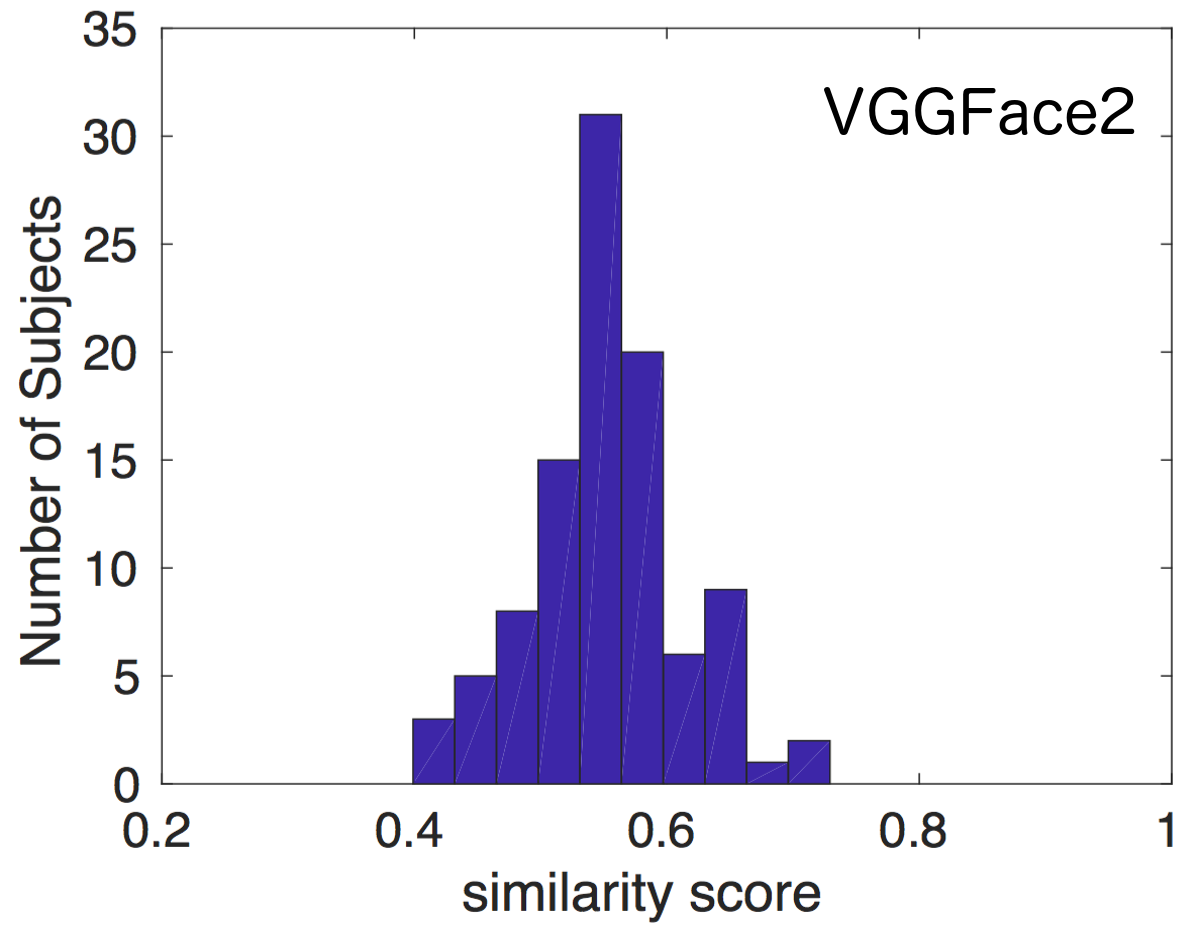}}
\end{center}
\caption{Histograms of similarity scores for young-to-mature matching for the models trained on different datasets.}
\label{age-analysis}
\end{figure*}

\begin{figure*}[ht!]
\captionsetup{font=small}
\begin{center}
\subfigure{\includegraphics[width=50mm]{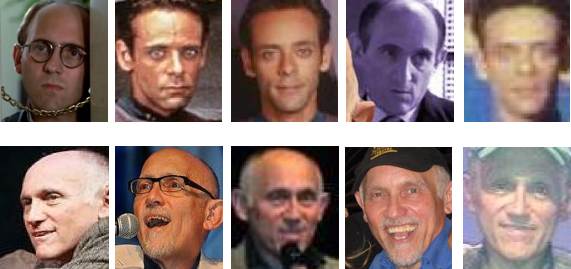}}  \hspace{20mm}
\subfigure{\includegraphics[width=50mm]{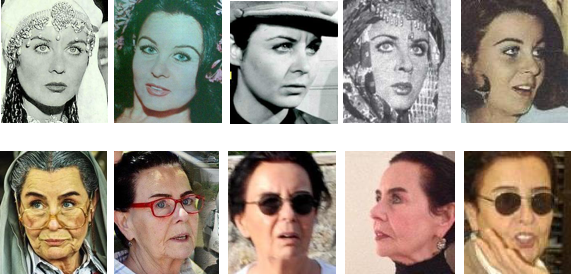}} 
\end{center}
\caption{Two example templates of young-to-mature matching. Left:  the similarity scores produced by VGGFace, MS1M, \NewVGGFace{} are $0.42$, $0.30$ and $0.58$, respectively; Right:  the scores are $0.43$, $0.41$ and $0.73$, respectively.}
\label{age-temp-analysis-samples}
\end{figure*}

%% file: texts/template_matches.tex
\begin{figure*}[htb!]
\captionsetup{font=small}
\begin{center}
\includegraphics[width=50mm]{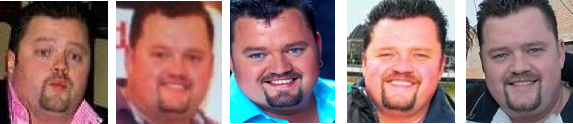} \hspace{20mm} 
\includegraphics[width=50mm]{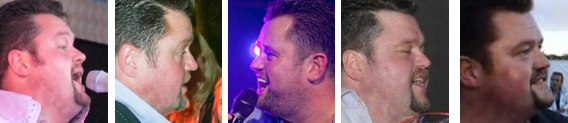} \\

\includegraphics[width=50mm]{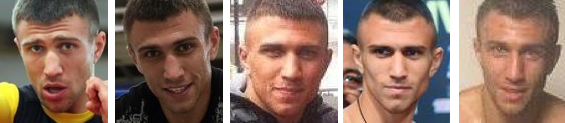} \hspace{20mm}  \includegraphics[width=50mm]{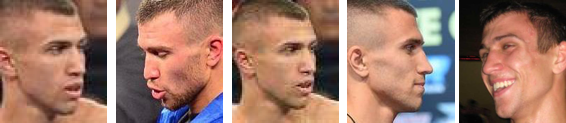} \\

\includegraphics[width=50mm]{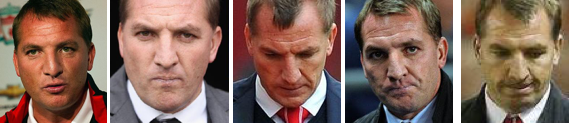} \hspace{20mm}  \includegraphics[width=50mm]{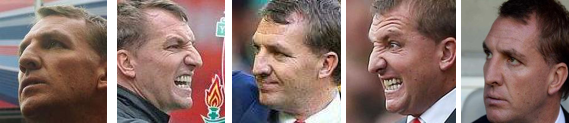} \\
\vspace{3mm}
\includegraphics[width=50mm]{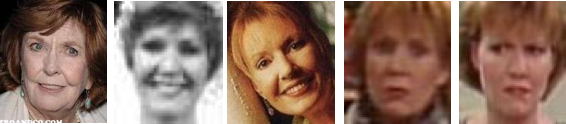} \hspace{20mm} \includegraphics[width=50mm]{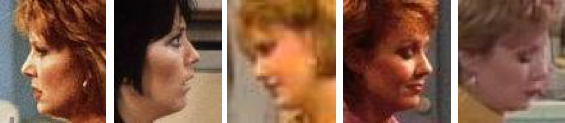}\\

\includegraphics[width=50mm]{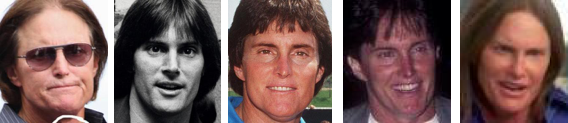} \hspace{20mm}  \includegraphics[width=50mm]{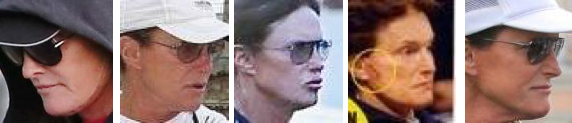} \\

\includegraphics[width=50mm]{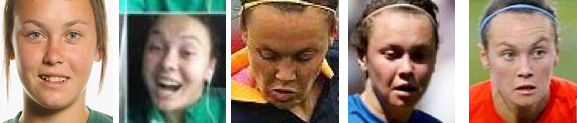} \hspace{20mm}  \includegraphics[width=50mm]{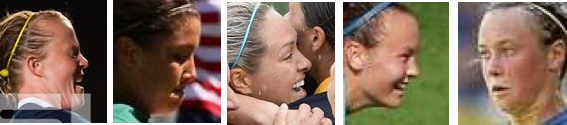} 
\end{center}
\caption{Top $3$ and bottom $3$ front-to-profile template matches sorted using the similarity score produced by
a ResNet  model trained on \NewVGGFace. The front template is on the left, and the profile template on the right in each row.}
\label{pose-temp-analysis-topsamples}
\end{figure*}

Figure~\ref{pose-temp-analysis-topsamples} shows the top $3$ and
bottom $3$ front-to-profile template matches sorted using the similarity
scores produced by the ResNet model trained on \NewVGGFace{}.  We can observe
that the model  gives high scores to front-to-profile templates
where there is little variation beyond pose; while it gives lower
scores to templates where many other variations exist, such as
expression and resolution.

\begin{figure*}[ht!]
\captionsetup{font=small}
\begin{center}
\includegraphics[width=50mm]{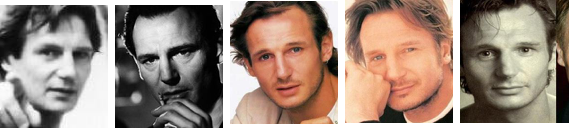} \hspace{20mm} 
\includegraphics[width=50mm]{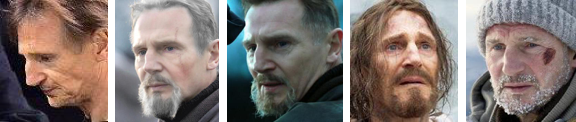} \\

\includegraphics[width=50mm]{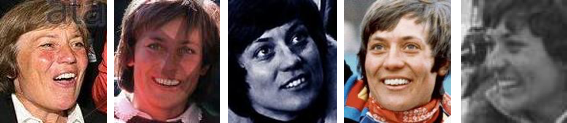} \hspace{20mm}  
\includegraphics[width=50mm]{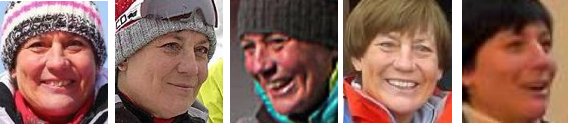} \\

\includegraphics[width=50mm]{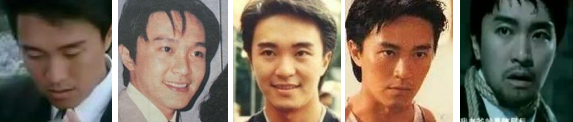} \hspace{20mm}  
\includegraphics[width=50mm]{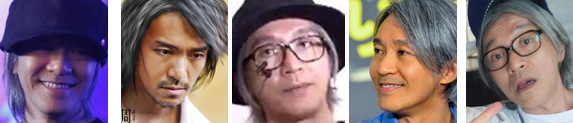} \\
\vspace{3mm}
\includegraphics[width=50mm]{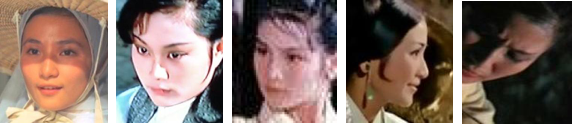} \hspace{20mm} \includegraphics[width=50mm]{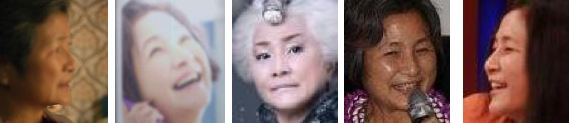} \\

\includegraphics[width=50mm]{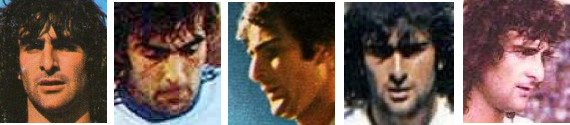} \hspace{20mm}  \includegraphics[width=50mm]{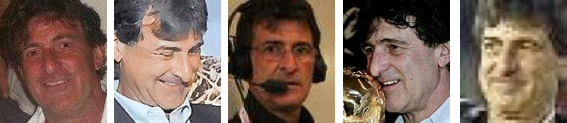} \\

\includegraphics[width=50mm]{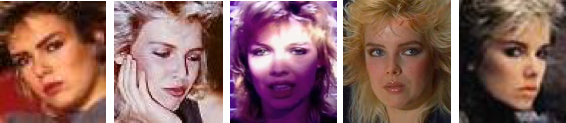} \hspace{20mm}  \includegraphics[width=50mm]{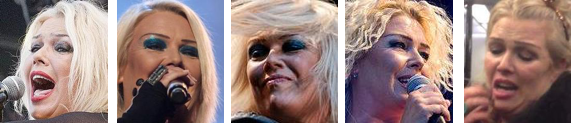} 
\end{center}
\caption{Top $3$ and bottom $3$ young-to-mature template matches sorted using the similarity score produced by 
a ResNet model trained on \NewVGGFace.  The young template is on the left, and the mature template on the right in each row.}
\label{age-temp-analysis-topsamples}
\end{figure*}

Figure~\ref{age-temp-analysis-topsamples} shows the top $3$ and bottom
$3$ young-to-mature template matches sorted using the similarity scores
produced by the ResNet  model trained on \NewVGGFace{}. It can be seen that
the model  gives low scores to young-to-mature templates where
variations such as pose and occlusion exist.

%% file: texts/IJBA_result.tex
\begin{table*}[ht]
\captionsetup{font=small}
\begin{center}{\scalebox{0.76}{
\begin{tabular}{|c|c|c|c|c|c|c|c|c|c|}
\hline
  Training dataset & Arch. & \multicolumn{3}{c|}{1:1 Verification TAR} &  \multicolumn{5}{c|}{1:N Identification TPIR}\\
 &  & FAR=$0.001$ & FAR=$0.01$ & FAR=$0.1$ & FPIR=$0.01$ & FPIR=$0.1$ & Rank-$1$ & Rank-$5$& Rank-$10$\\
\hline
VGGFace~\cite{Parkhi15}&  ResNet-50 & $0.620 \pm 0.043$ & $0.834 \pm 0.021$ & $0.954 \pm 0.005$ &  $0.454 \pm 0.058$ & $0.748 \pm 0.024$ & $0.925 \pm 0.008$ &  $0.972 \pm 0.005$ & $0.983 \pm 0.003$ \\
MS$1$M~\cite{guo2016ms} & ResNet-50 & $0.851 \pm 0.030$ & $0.939 \pm 0.013$ & $0.980 \pm 0.003$ & $0.807 \pm 0.041$ & $0.920 \pm 0.012$ & $0.961 \pm 0.006$ & $0.982 \pm 0.004$ & $0.990 \pm 0.002$\\
\NewVGGFace &  ResNet-50 & $0.895 \pm 0.019$ & $0.950 \pm 0.005$& $0.980 \pm 0.003$ &  $0.844 \pm 0.035$ & $0.924 \pm 0.006$ & $0.976 \pm 0.004$ &  $0.992 \pm 0.002$ & $0.995 \pm 0.001$ \\
\NewVGGFace\_ft &  ResNet-50 & $0.908 \pm 0.017$ & $0.957 \pm 0.007$ & $0.986 \pm 0.002$ &  $0.861 \pm 0.027$ & $0.936 \pm 0.007$ & $0.978 \pm 0.005$ & $0.992 \pm 0.003$ & $0.995 \pm 0.001$\\
\NewVGGFace & SENet & $0.904 \pm 0.020$ & $0.958 \pm 0.004$ & $0.985 \pm 0.002$ &$0.847 \pm 0.051$ & $0.930 \pm 0.007$ & $0.981\pm 0.003$ & $\mathbf{0.994 \pm 0.002}$  & $\mathbf{0.996 \pm 0.001}$ \\
\NewVGGFace\_ft & SENet & $\mathbf{0.921\pm 0.014}$ & $\mathbf{0.968 \pm 0.006}$ & $\mathbf{0.990 \pm 0.002}$ & $\mathbf{0.883 \pm 0.038}$ & $\mathbf{0.946 \pm 0.004}$ & $\mathbf{0.982 \pm 0.004}$ & $0.993 \pm 0.002$ & $0.994 \pm 0.001$\\
\hline\hline
Crosswhite {\it et al.}~\cite{crosswhite2017template} & - & $0.836 \pm 0.027$ & $0.939 \pm 0.013$& $0.979 \pm 0.004$ & $0.774 \pm 0.049$ & $0.882 \pm 0.016$ & $0.928 \pm 0.010$ & $0.977 \pm 0.004$ & $0.986 \pm 0.003$ \\ \hline
Sohn {\it et al.}~\cite{sohn2017unsupervised} &  - & $0.649 \pm 0.022$ & $0.864 \pm 0.007$& $0.970 \pm 0.001$ & - & - & $0.895 \pm 0.003$ & $0.957 \pm 0.002$ & $0.968 \pm 0.002$ \\ \hline
Bansal{it et al.}~\cite{bansal2017s} & - & $0.730^{\dagger}$ & $0.874$ & $ 0.960^{\dagger}$ & - & - & - & - & - \\ \hline
Yang {\it et al.}~\cite{yang2017neural} & - & $0.881 \pm 0.011$ & $0.941 \pm 0.008$ & $0.978 \pm 0.003$ & $0.817 \pm 0.041$ & $0.917 \pm 0.009$ & $0.958 \pm 0.005$ & $0.980 \pm 0.005$ & $0.986 \pm 0.003$ \\
 \hline
\end{tabular}}}
\end{center}
\vspace{-1.5mm}
\caption{Performance evaluation on the IJB-A dataset. A  higher value is  better. The values with ${\dagger}$ are read from~\cite{bansal2017s}.}
\label{jiba-evaluation-table}
\end{table*}

\begin{figure*}[ht]
\captionsetup{font=small}
\begin{center}
\subfigure{\includegraphics[width=55mm]{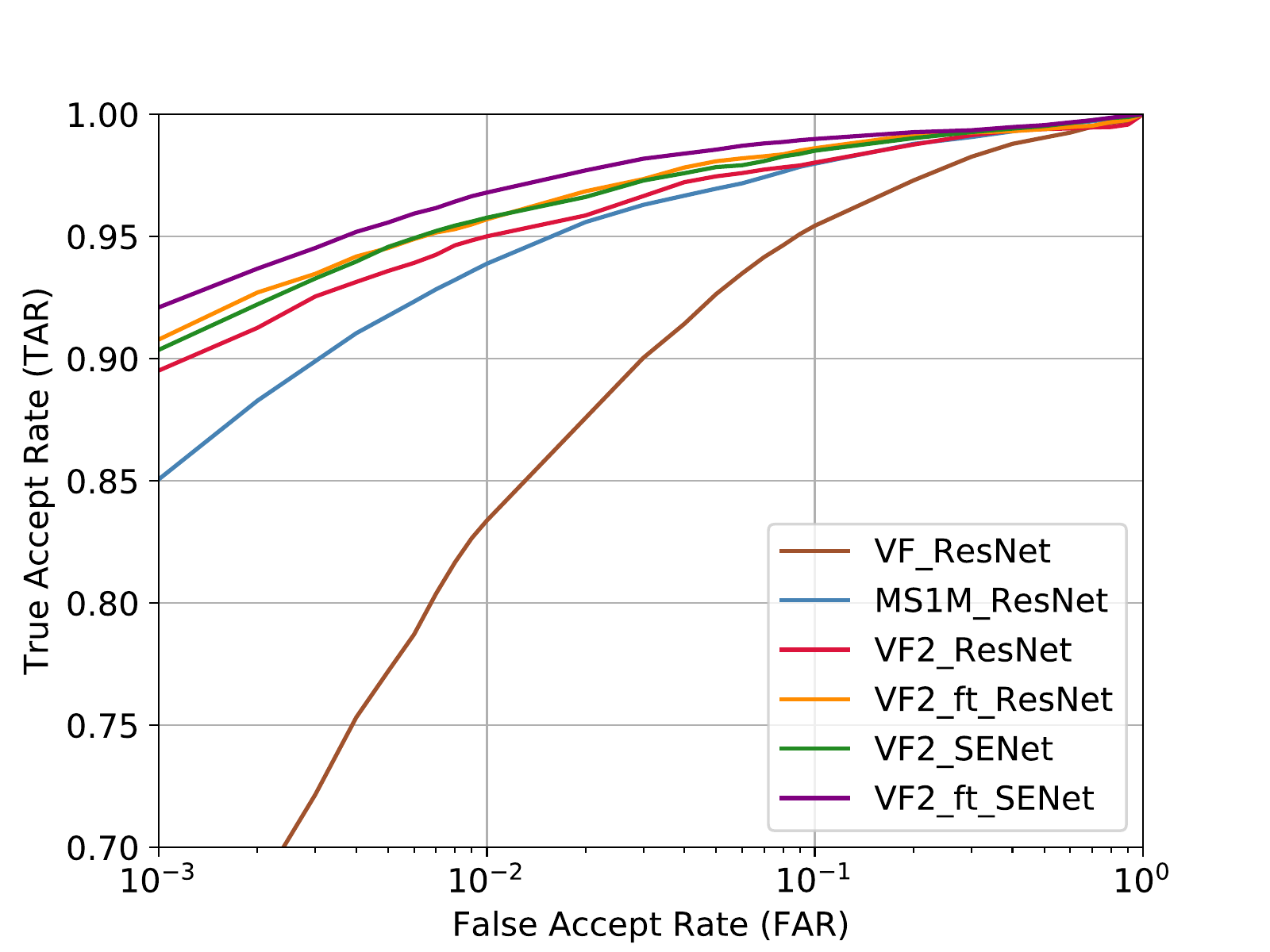}} \hspace{4mm}
 \subfigure{\includegraphics[width=55mm]{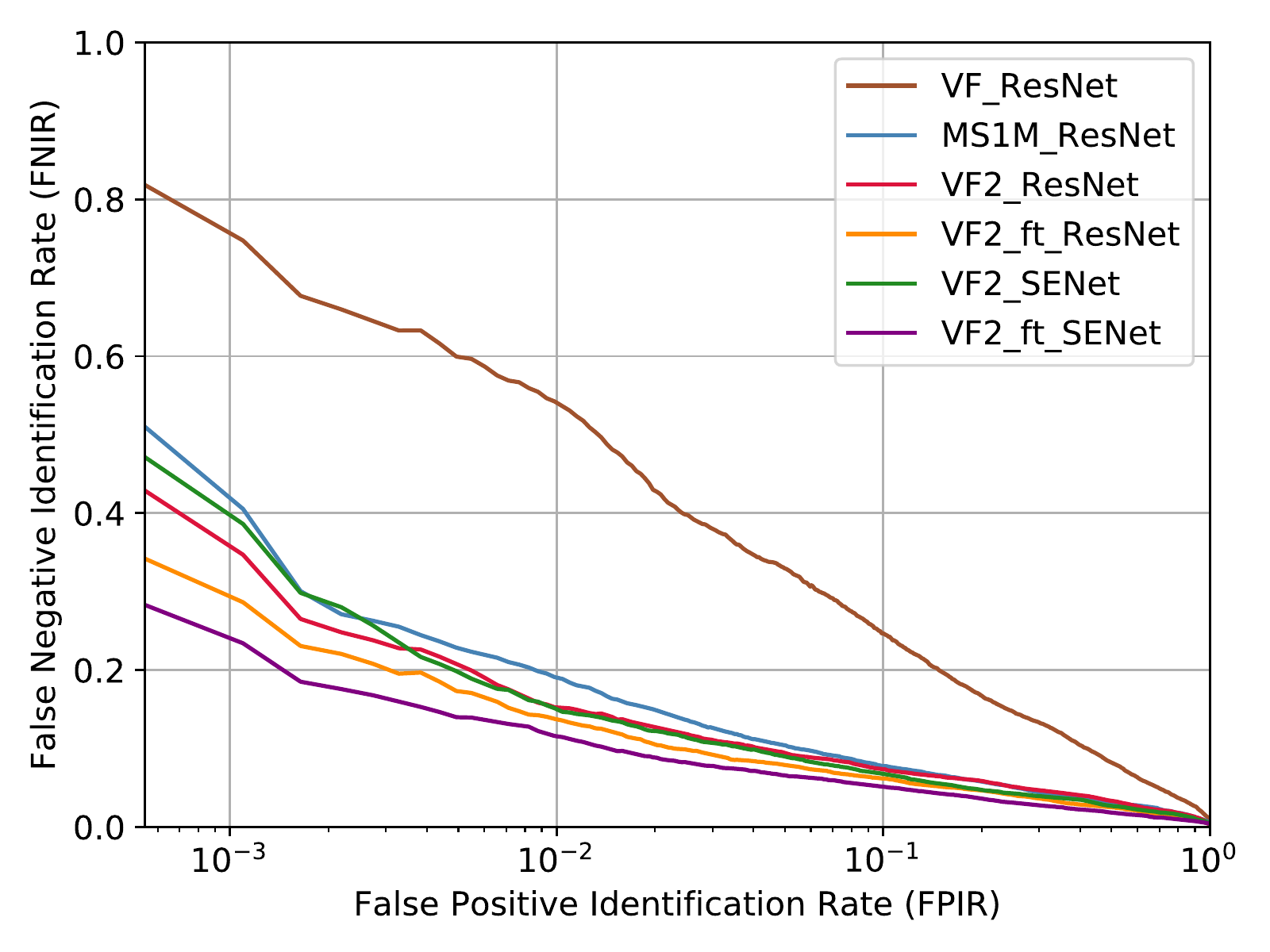}}  \hspace{4.2mm}
 \subfigure{\includegraphics[width=55mm]{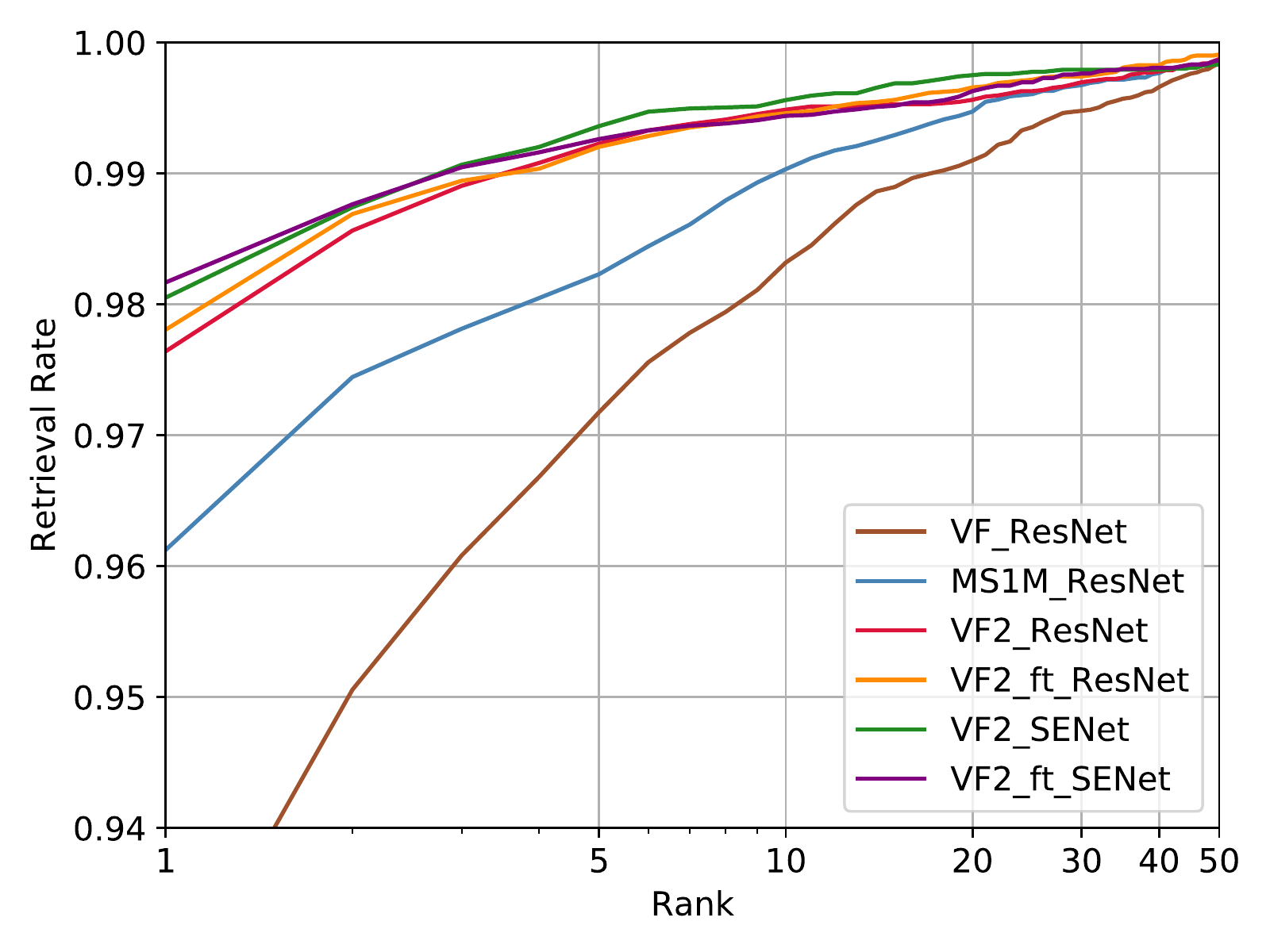}} 
\end{center}
\vspace{-2mm}
\caption{Results on the IJB-A dataset (average over 10 splits). 
Left: ROC (higher is better);  
Middle: DET (lower is better);  
Right: CMC (higher is better).}
\label{ijba-curves}
\end{figure*}

%% file: texts/IJBB_result.tex
\begin{table*}[ht!]
\captionsetup{font=small}
\begin{center}{\scalebox{0.72}{
\begin{tabular}{|c|c|c|c|c|c|c|c|c|c|c|}
\hline
 Training dataset & Arch. & \multicolumn{4}{c|}{1:1 Verification TAR} &  \multicolumn{5}{c|}{1:N Identification TPIR}\\
 &  & FAR=$1E-5$ & FAR=$1E-4$ & FAR=$1E-3$ & FAR=$1E-2$ & FPIR=$0.01$ & FPIR=$0.1$ & Rank-$1$ & Rank-$5$& Rank-$10$\\
\hline
VGGFace~\cite{Parkhi15}& ResNet-50 & $0.342$ & $0.535$ & $0.711$ &  $0.850$ &$0.429 \pm 0.024$& $0.635 \pm 0.015$ &$0.752 \pm 0.038$& $0.843 \pm 0.032$& $0.874 \pm 0.026$\\
MS$1$M~\cite{guo2016ms} & ResNet-50 & $0.548$ & $0.743$ & $0.857$ & $0.935$ & $0.662 \pm 0.036$& $0.810 \pm 0.028$ & $0.865 \pm 0.053$ & $0.917 \pm 0.032$ & $0.936 \pm 0.024$\\
\NewVGGFace & ResNet-50 & $0.647$ & $0.784$ & $0.878$ &  $0.938$ & $0.701 \pm 0.038$& $0.824 \pm 0.034$ & $0.886 \pm 0.032$ & $0.936 \pm 0.019$ & $0.953 \pm 0.013$\\
\NewVGGFace\_ft &  ResNet-50 & $0.671$ & $0.804$ & $0.891$ &  $0.947$ & $0.702 \pm 0.041$ & $0.843 \pm 0.032$ & $0.894 \pm 0.039$ & $0.940 \pm 0.022$ & $0.954 \pm 0.016$\\
\NewVGGFace & SENet & $0.671$ & $0.800$ & $0.888$ & $0.949$ & $0.706 \pm 0.047 $ & $0.839 \pm 0.035$ & $0.901 \pm 0.030$ & $0.945 \pm 0.016$ & $0.958 \pm 0.010$ \\
\NewVGGFace\_ft & SENet & $\mathbf{0.705}$ & $\mathbf{0.831}$ & $\mathbf{0.908}$ & $\mathbf{0.956}$ & $\mathbf{0.743 \pm 0.037}$ & $\mathbf{0.863 \pm 0.032}$ & $\mathbf{0.902 \pm 0.036}$ & $\mathbf{0.946 \pm 0.022}$ & $\mathbf{0.959 \pm 0.015}$ \\
 \hline \hline
Whitelam {\it et al.}~\cite{whitelam2017iarpa}& - & $0.350$ & $0.540$ & $0.700$ & $0.840$ & $0.420$ & $0.640$ & $0.790$ & $0.850$ &$0.900$ \\
\hline
\end{tabular}}}
\end{center}
\vspace{-1.5mm}
\caption{
Performance evaluation on the IJB-B dataset. A higher value is
better. The results of~\cite{whitelam2017iarpa} are read from the
curves reported in the paper. Note, \cite{whitelam2017iarpa} has a different
evaluation for the verification protocol where  pairs
generated from different galleries are evaluated separately and
averaged to get the final results.}
\label{jibb-evaluation-table}
\end{table*}

\begin{figure*}[ht!]
\captionsetup{font=small}
\begin{center}
\subfigure{\includegraphics[width=55mm]{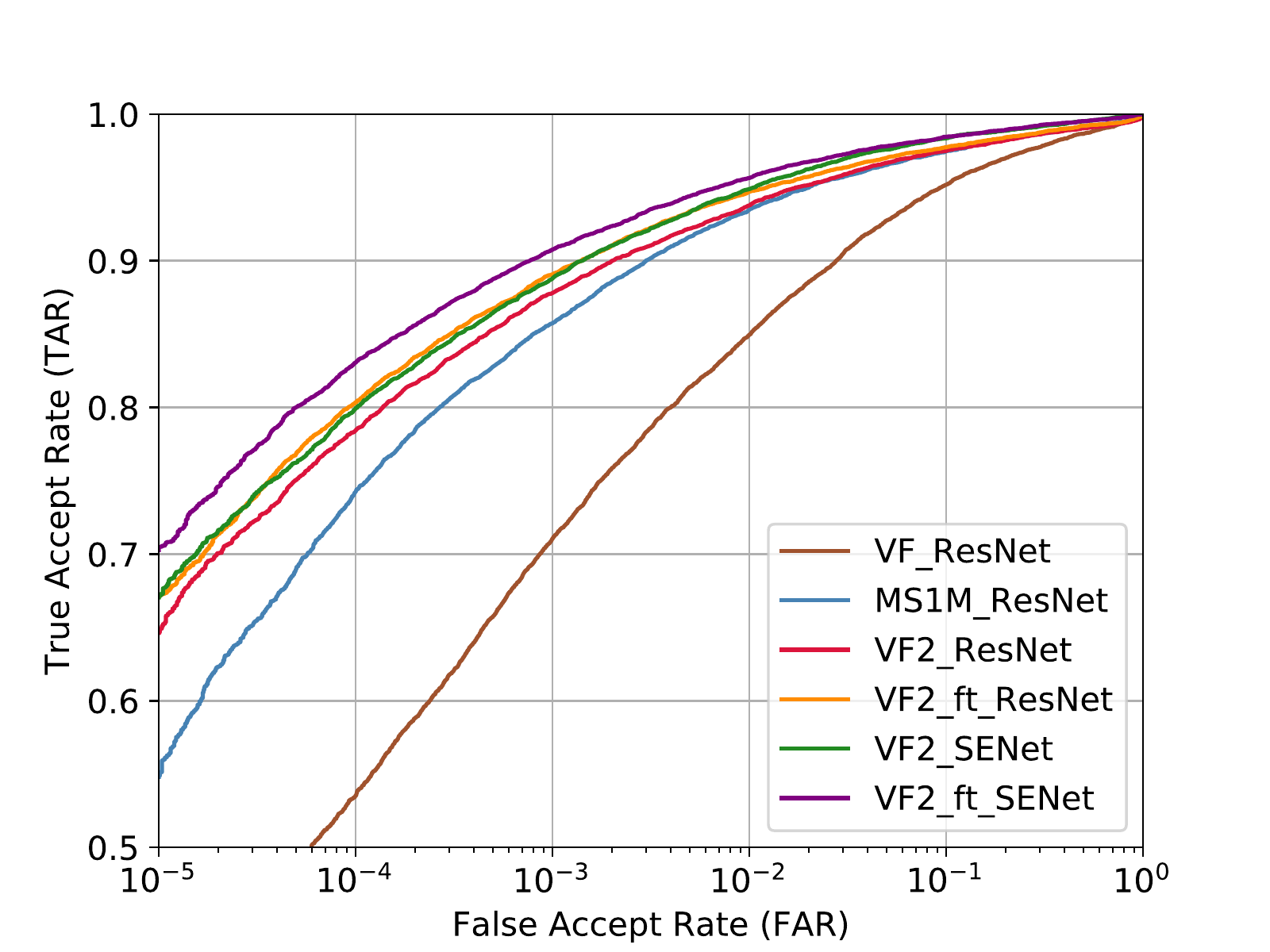}} \hspace{4mm}
\subfigure{\includegraphics[width=55mm]{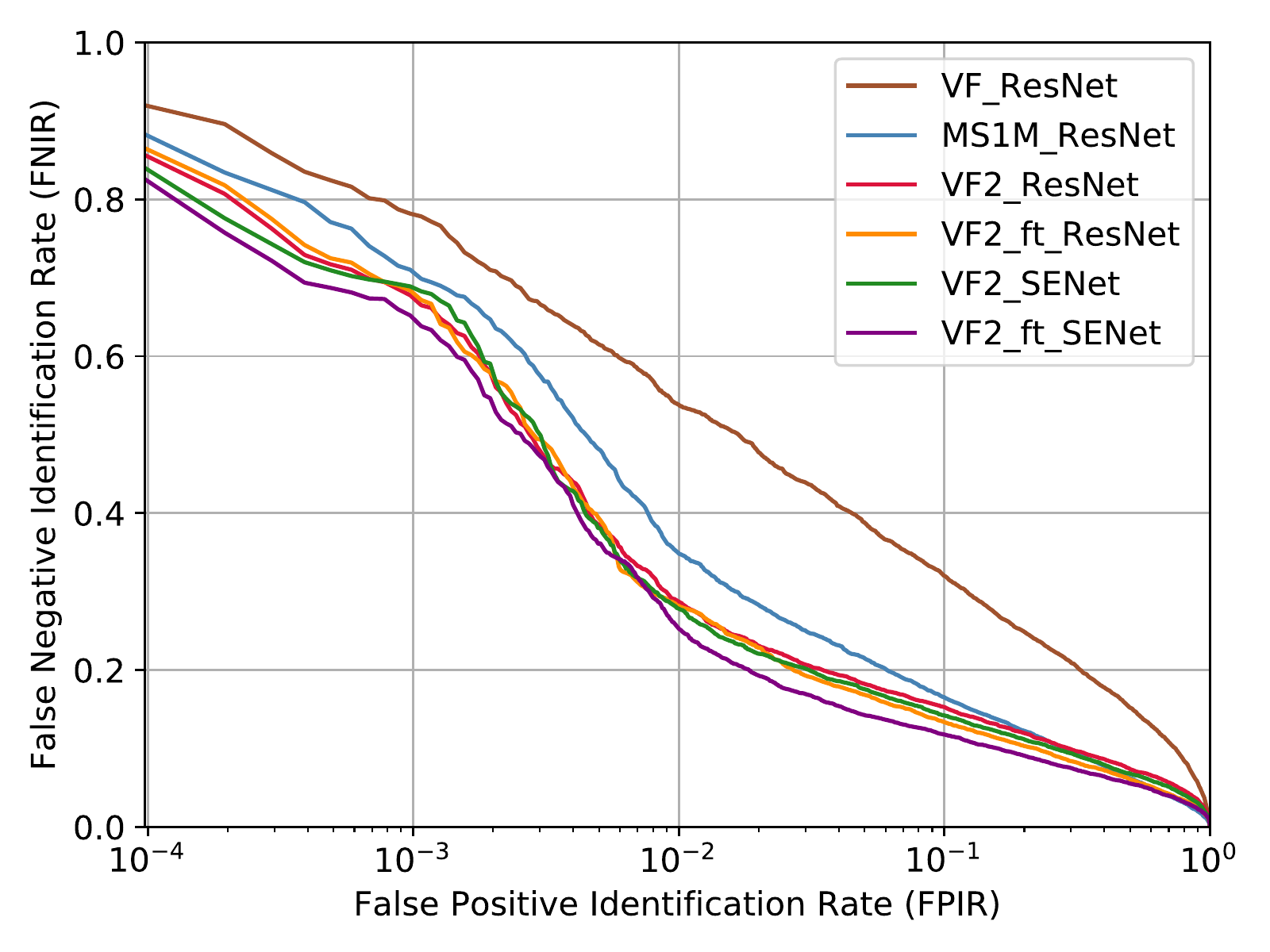}}\hspace{4.2mm}
\subfigure{\includegraphics[width=55mm]{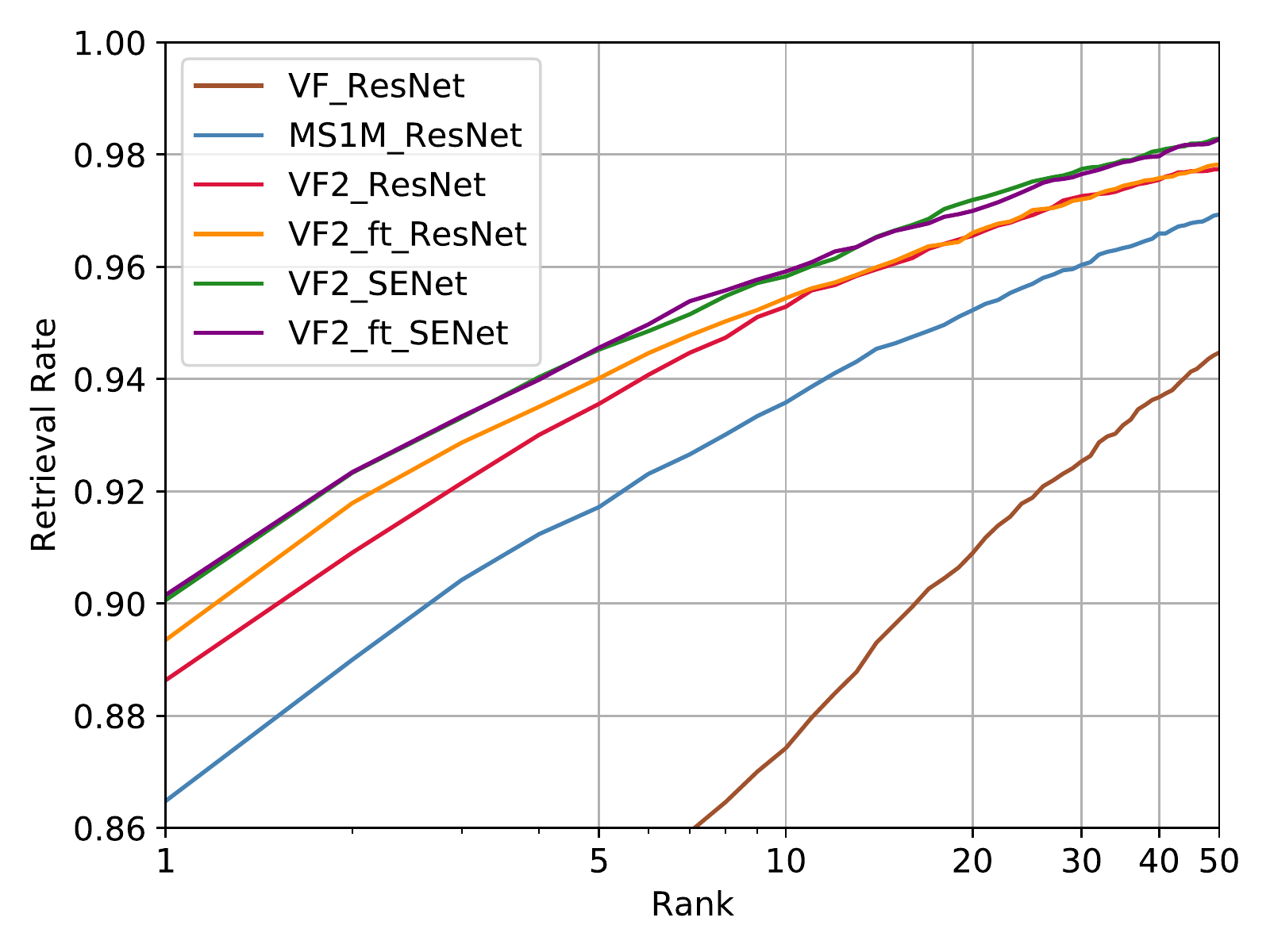}}
\end{center}
\vspace{-2mm}
\caption{Results on the IJB-B dataset across gallery sets S1 and S2. \\
Left: ROC (higher is better);  
Middle: DET (lower is better);  
Right: CMC (higher is better).}
\label{ijbb-curves}
\end{figure*}

%% file: texts/IJBC_result.tex
\begin{table*}[ht!]
\captionsetup{font=small}
\begin{center}{\scalebox{0.72}{
\begin{tabular}{|c|c|c|c|c|c|c|c|c|c|c|}
\hline
 Training dataset & Arch. & \multicolumn{4}{c|}{1:1 Verification TAR} &  \multicolumn{5}{c|}{1:N Identification TPIR}\\
 &  & FAR=$1E-5$ & FAR=$1E-4$ & FAR=$1E-3$ & FAR=$1E-2$ & FPIR=$0.01$ & FPIR=$0.1$ & Rank-$1$ & Rank-$5$& Rank-$10$\\
\hline
\NewVGGFace & ResNet-50 & $0.734$ & $0.825$ & $0.900$ &  $0.950$ & $0.735 \pm 0.022$& $0.830 \pm 0.021$ & $0.898 \pm 0.017$ & $0.939 \pm 0.013$ & $0.953 \pm 0.009$\\
\NewVGGFace\_ft &  ResNet-50 & $0.749$ & $0.846$ & $0.913$ &  $0.958$ & $0.749 \pm 0.021$ & $0.849 \pm 0.018$ & $0.908 \pm 0.022$ & $0.946 \pm 0.014$ & $0.958 \pm 0.011$\\
\NewVGGFace & SENet & $0.747$ & $0.840$ & $0.910$ & $0.960$ & $0.746 \pm 0.018 $ & $0.842 \pm 0.022$ & $0.912 \pm 0.017$ & $0.949 \pm 0.010$ & $0.962 \pm 0.007$ \\
\NewVGGFace\_ft & SENet & $\mathbf{0.768}$ & $\mathbf{0.862}$ & $\mathbf{0.927}$ & $\mathbf{0.967}$ & $\mathbf{0.763 \pm 0.018}$ & $\mathbf{0.865 \pm 0.018}$ & $\mathbf{0.914 \pm 0.020}$ & $\mathbf{0.951 \pm 0.013}$ & $\mathbf{0.961 \pm 0.010}$ \\
 \hline \hline
Maze {\it et al.}~\cite{Maze2018}& & $0.600$ & $0.750$ & $0.860$ & $0.950$ & $0.450$ & $0.620$ & $0.790$ & $0.870$ &$0.900$ \\
\hline
\end{tabular}}}
\end{center}
\vspace{-1.5mm}
\caption{
Performance evaluation on the IJB-C dataset. A higher value is
better. The results of~\cite{Maze2018} are read from the
curves reported in the paper. Note, \cite{Maze2018} has a different
evaluation for the verification protocol where  pairs
generated from different galleries are evaluated separately and
averaged to get the final results.}
\label{jibc-evaluation-table}
\end{table*}

\begin{figure*}[ht!]
\captionsetup{font=small}
\begin{center}
\subfigure{\includegraphics[width=58mm]{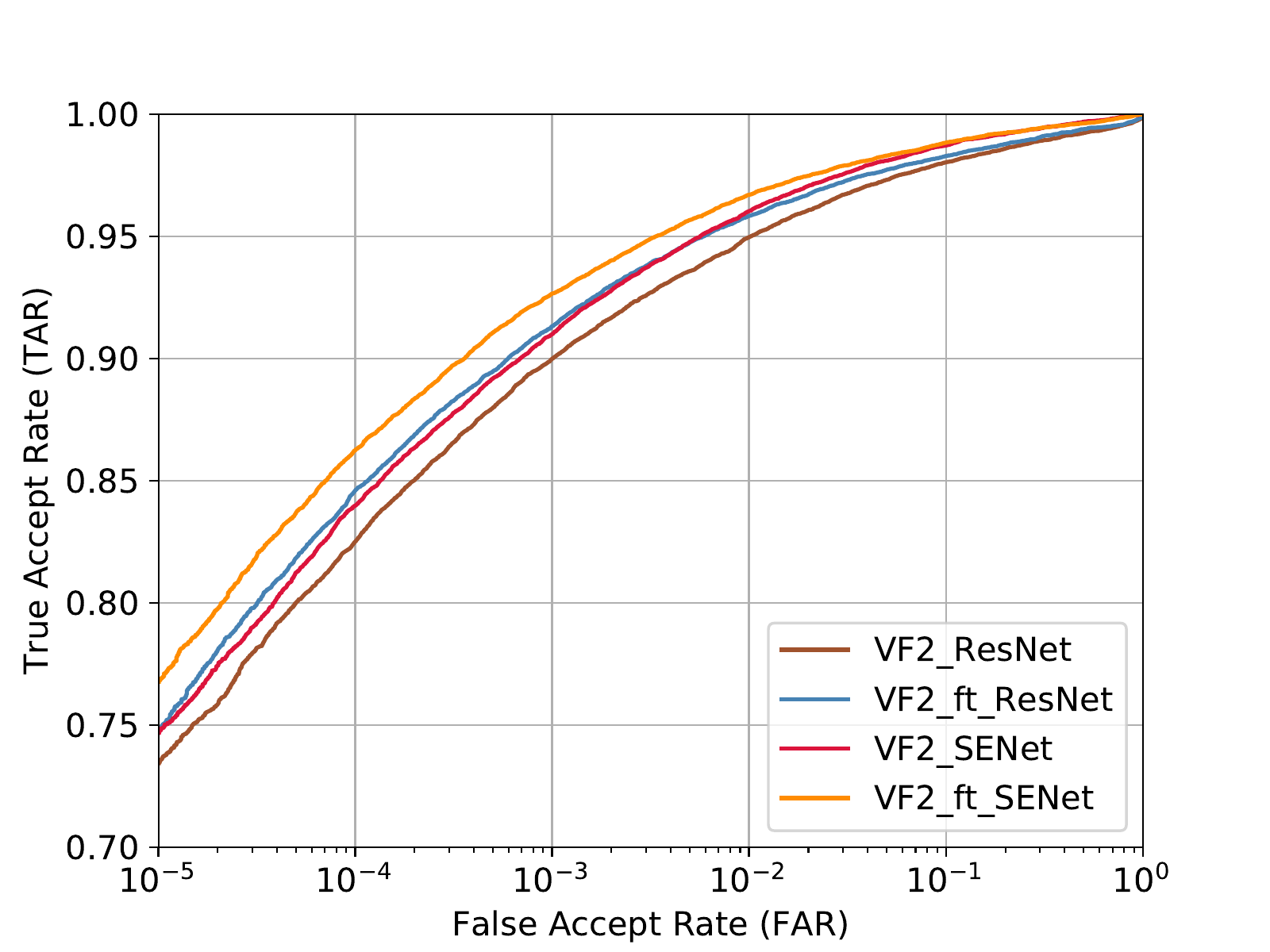}} \hspace{3mm}
\subfigure{\includegraphics[width=54.5mm]{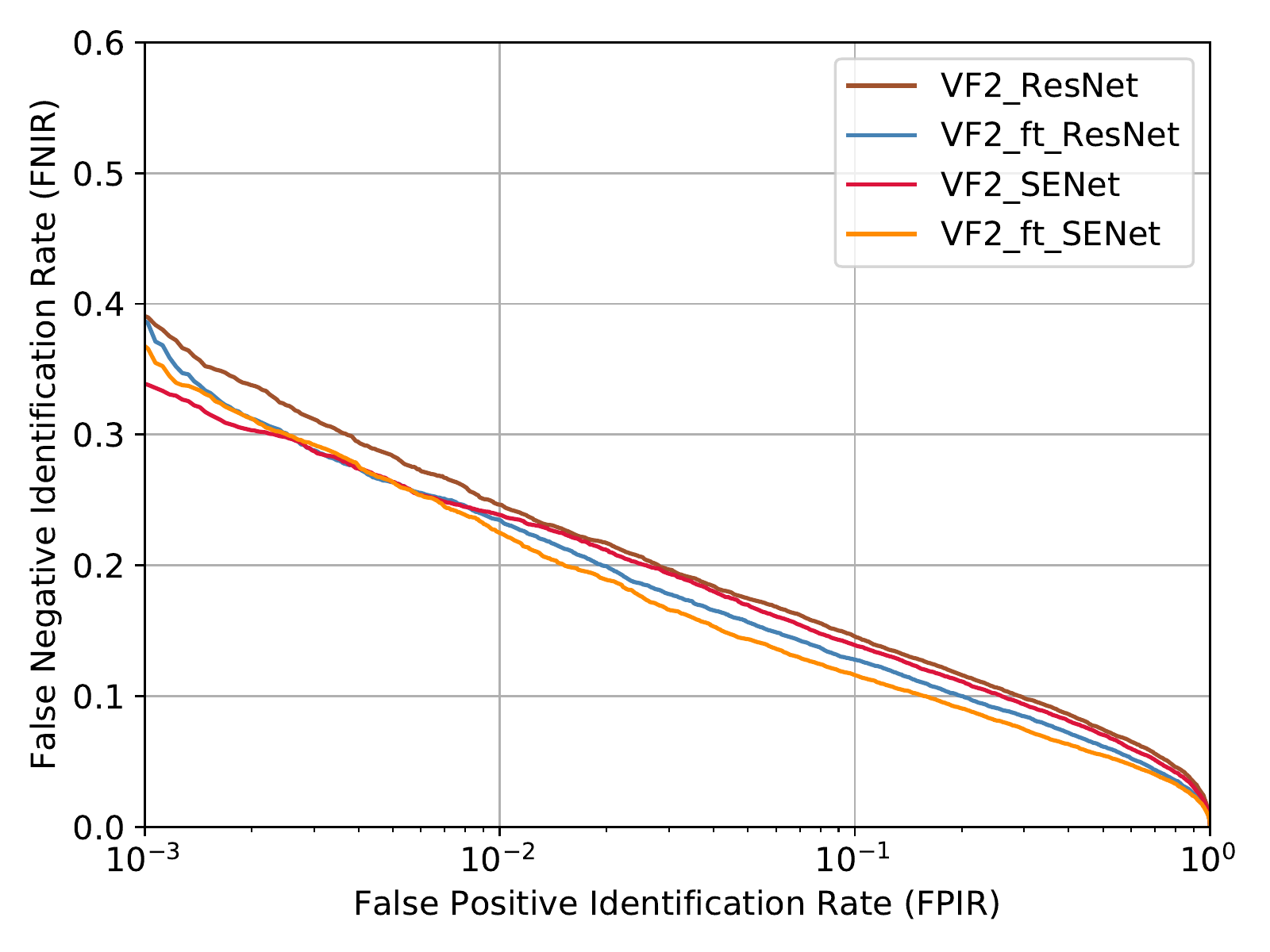}}\hspace{3mm}
\subfigure{\includegraphics[width=54.5mm]{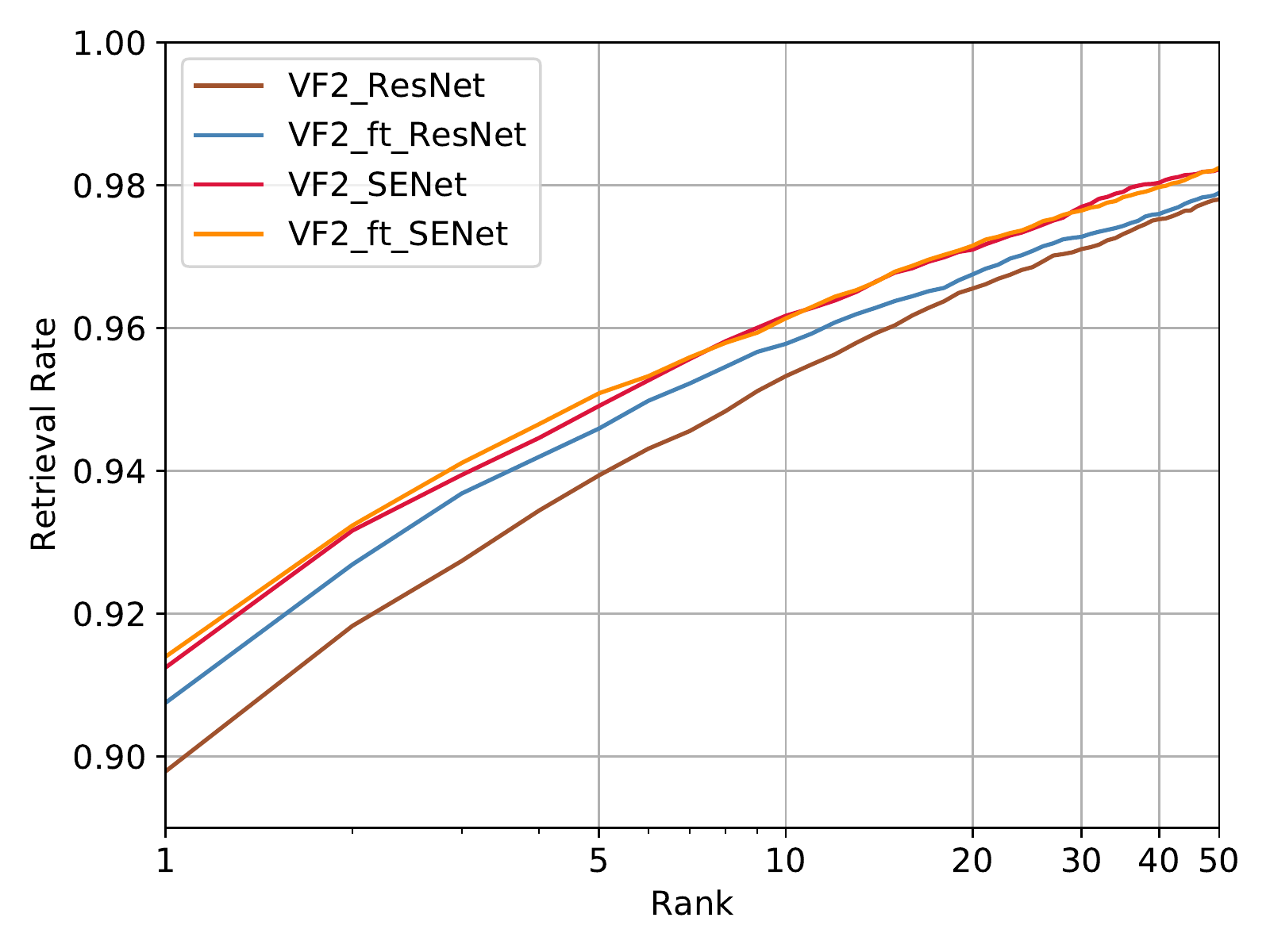}}
\end{center}
\vspace{-2mm}
\caption{Results on the IJB-C dataset across gallery sets G1 and G2. \\
Left: ROC (higher is better);  
Middle: DET (lower is better);  
Right: CMC (higher is better).}
\label{ijbc-curves}
\end{figure*}

%% file: texts/conclusion.tex
In this work, we have proposed a pipeline for collecting a high-quality dataset,  \NewVGGFace{},  
with a wide range of pose and age.
Furthermore, we demonstrate that deep models (ResNet-50 and SENet) trained on \NewVGGFace{}, 
achieve state-of-the-art performance on the IJB-A, IJB-B and IJB-C  benchmarks. The dataset and models are 
available at \url{https://www.robots.ox.ac.uk/~vgg/data/vgg_face2/}.